\documentclass[runningheads]{llncs}
\usepackage{graphicx}
\usepackage[caption=false]{subfig}
\usepackage{comment}
\usepackage{amsmath,amssymb} 
\usepackage{rotating}
\usepackage{color}
\usepackage{multirow}
\begin{document}
\pagestyle{headings}
\mainmatter
\def\ECCVSubNumber{2142}  

\title{The Devil is in the Details: Self-Supervised Attention for Vehicle Re-Identification} 

\titlerunning{Self-Supervised Attention for Vehicle Re-ID}
%
\author{Pirazh Khorramshahi\thanks{The first two authors equally contributed to this work.}\inst{1}
\and
$\text{Neehar Peri}^\star$\inst{1}
\and
Jun-cheng Chen\inst{2}
\and \\ Rama Chellappa \inst{1}}
\authorrunning{P. Khorramshahi et al.}
%
\institute{Center for Automation Research, UMIACS, and the Department of Electrical and Computer Engineering, University of Maryland, College Park\\ \and 
Research Center for Information Technology Innovation, Academia Sinica}


\maketitle

\begin{abstract}
In recent years, the research community has approached the problem of vehicle re-identification (re-id) with attention-based models, specifically focusing on regions of a vehicle containing discriminative information. These re-id methods rely on expensive key-point labels, part annotations, and additional attributes including vehicle make, model, and color. Given the large number of vehicle re-id datasets with various levels of annotations, strongly-supervised methods are unable to scale across different domains. In this paper, we present Self-supervised Attention for Vehicle Re-identification (SAVER), a novel approach to effectively learn vehicle-specific discriminative features. Through extensive experimentation, we show that SAVER improves upon the state-of-the-art on challenging VeRi, VehicleID, Vehicle-1M and VERI-Wild datasets. 

\end{abstract}

\section{Introduction}
 Re-identification (re-id), the task of identifying all images of a specific object ID in a gallery, has been recently revolutionized with the advancement of Deep Convolutional Neural Networks (DCNNs). This revolution is most notable in the area of person re-id. Lou \emph{et al.} \cite{luo2019bag} recently developed a strong baseline method that supersedes state-of-the-art person re-id methods by a large margin, using an empirically derived ``Bag of Tricks'' to improve the discriminative capacity of DCNNs. This has created a unique opportunity for the research community to develop innovative yet simple methods to push the boundaries of object re-id. 

Specifically, vehicle re-id has great potential in intelligent transportation applications. However, the task of vehicle re-id is particularly challenging since vehicles with different identities can be of the same make, model and color. Moreover, the appearance of a vehicle varies significantly across different viewpoints. Therefore, recent DCNN-based re-id methods focus attention on discriminative regions to improve robustness to orientation and occlusion. To this end, many high performing re-id approaches rely on additional annotations for local regions that have been shown to carry identity-dependent information, \emph{i.e.} key-points \cite{wang2017orientation,Khorramshahi_2019_ICCV,Khorramshahi_2019_CVPR_Workshops} and parts bounding boxes \cite{he2019part,zhang2019part} in addition to the ID of the objects of interest. These extra annotations help DCNNs jointly learn improved global and local representations and significantly boost performance \cite{Khorramshahi_2019_ICCV,zheng2019attributes} at the cost of increased complexity. Despite providing considerable benefit, gathering costly annotations such as key-point and part locations cannot be scaled to the growing size of vehicle re-id datasets. As manufacturers change the design of their vehicles, the research community has the burdensome task of annotating new vehicle models.
\begin{figure}[t]
    \centering
    \subfloat[]{\includegraphics[width=0.14\textwidth]{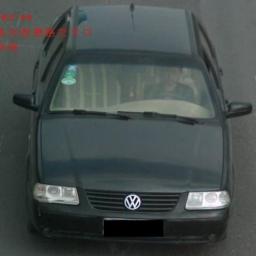}}\quad
    \subfloat[]{\includegraphics[width=0.14\textwidth]{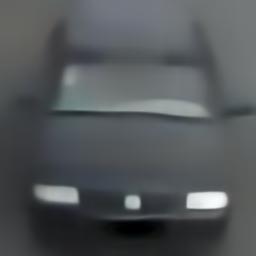}}\quad
    \subfloat[]{\includegraphics[width=0.14\textwidth]{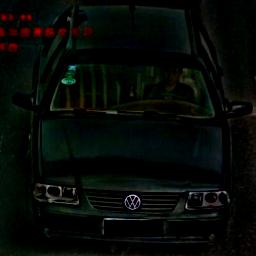}}\quad
    \subfloat[]{\includegraphics[width=0.14\textwidth]{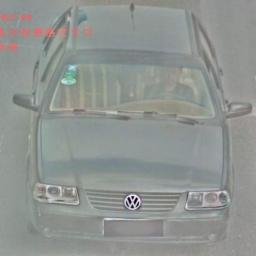}}\\
    \subfloat[]{\includegraphics[width=0.14\textwidth]{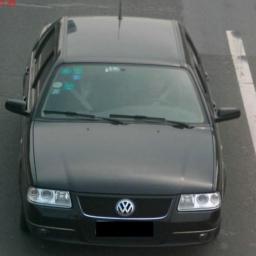}}\quad
    \subfloat[]{\includegraphics[width=0.14\textwidth]{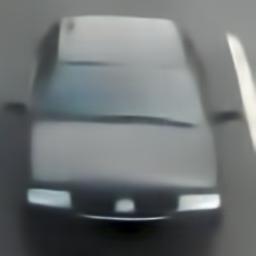}}\quad
    \subfloat[]{\includegraphics[width=0.14\textwidth]{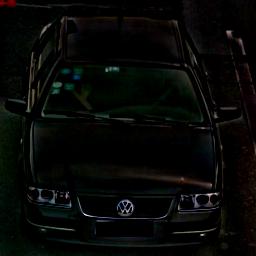}}
    \quad
    \subfloat[]{\includegraphics[width=0.14\textwidth]{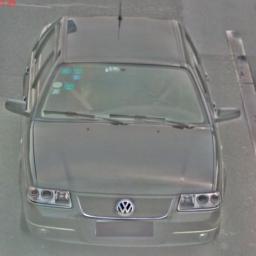}}
    \caption{Vehicle image decomposition into coarse reconstruction and residual images, left-most column (a,e):  vehicle image, second column (b,f): coarse reconstruction, third column (c,g): residual, right-most column (d,h): normalized residual (for the sake of visualization). Despite having the same coarse reconstruction, both vehicles have different residuals highlighting key areas, \emph{e.g.}, the windshield stickers, bumper design.}
    \label{fig:motivation}
\end{figure}
In an effort to re-design the vehicle re-id pipeline without the need for expensive annotations, we propose SAVER to automatically highlight salient regions in a vehicle image. These vehicle-specific salient regions carry critical details that are essential for distinguishing two visually similar vehicles. Specifically, we design a Variational Auto-Encoder (VAE) \cite{kingmaVAE} to generate a vehicle image template that is free from manufacturer logos, windshield stickers, wheel patterns, and grill, bumper and head/tail light designs. 
By obtaining this coarse reconstruction and its pixel-wise difference from the original image, we construct \textbf{residual} image. This residual contains crucial details required for re-id, and acts as a pseudo-saliency or pseudo-attention map highlighting discriminative regions in an image. Fig. \ref{fig:motivation} shows how the residual map highlights valuable fine-grained details needed for re-identification between two visually similar vehicles. 

The rest of the paper is organized as follows. In section \ref{sec:related}, we briefly review recent works in vehicle re-id. The detailed architecture of each step in the proposed approach is discussed in section \ref{sec:proposed}. Through extensive experimentation in section \ref{sec:experiments}, we show the effectiveness of our approach on multiple challenging vehicle re-id benchmarks  \cite{yan2017_PKUVD,liu2016_VehicleID,Guo2018_Vehicle1M,Yihang2019_VeRi_Wild,Xinchen2016_VeRi776}, obtaining state-of-the-art results. Finally, in section \ref{sec:ablation} we validate our design choices.

\section{Related Works}
\label{sec:related}
Learning robust and discriminative vehicle representations that adapt to large viewpoint variations across multiple cameras, illumination and occlusion is essential for re-id. Due to a large volume of literature, we briefly review recent works on vehicle re-identification.

With recent breakthroughs due to deep learning, we can easily learn discriminative embeddings for vehicles by feeding images from large-scale vehicle datasets, such as  
VehicleID, VeRi, VERI-Wild, Vehicle-1M, PKU VD1$\&$VD2~\cite{yan2017_PKUVD}, CompCars~\cite{Linjie2015_CompCars}, and CityFlow~\cite{Tang2019_CityFlow}, to train a DCNN that is later used as the feature extractor for re-id. However, for vehicles of the same make, model, and color, this global deep representation usually fails to discriminate between two similar-looking vehicles. To address this issue, several auxiliary features and strategies are proposed to enhance the learned global appearance representation. Cui \emph{et al.}~\cite{cui2017vehicle} fuse features from various DCNNs trained with different objectives. Suprem \emph{et al.}~\cite{suprem2019robust} propose the use of an ensemble of re-id models for vehicle identity and attributes for robust matching. \cite{wang2017orientation,liu2018ram,zhang2019part,he2019part,Khorramshahi_2019_ICCV} propose learning enhanced representation by fusing global features with auxiliary 
local representations learned from prominent vehicle parts and regions, \emph{e.g.,} headlights, mirrors. Furthermore, Peng \emph{et al.} ~\cite{peng2019eliminating} leverage an image-to-image translation model to reduce cross-camera bias for vehicle images from different cameras before learning auxiliary local representation.
Zhou \emph{et al.}~\cite{Zhou_2018_CVPR} learn vehicle representation via viewpoint-aware attention. Similarly, \cite{zheng2019attributes,qian2019stripe} leverage attention guided by vehicle attribute classification, \emph{e.g.}, color and vehicle type, to learn attribute-based auxiliary features to enhance the global representation. Metric learning is another popular approach to make representations more discriminative. \cite{zhang2017improving,bai2018group,chu2019vehicle,kuma2019vehicle} propose various triplet losses to carefully select hard triplets across different viewpoints and vehicles to learn an improved appearance-robust representation.

Alternatively, to augment training data for more robust training, \cite{yao2019simulating} adopts a graphic engine and \cite{wu2018joint,tang2019pamtri} use generative adversarial networks (GANs) to synthesize vehicle images with diverse orientations, appearance variations, and other attributes. \cite{liu2016deep,liu2017provid,shen2017learning,tan2019multi,hsu2019multi,lv2019vehicle,huang2019multi} propose methods for improving the matching performance by also making use of spatio-temporal and multi-modal information, such as visual features, license plates, inter-camera vehicle trajectories, camera locations, and time stamps. 

In contrast with prior methods, SAVER benefits from self-supervised attention generation and does not assume any access to extra annotations, attributes, spatio-temporal and multi-modal information.

\section{Self-Supervised Attention for Vehicle Re-identification}
\label{sec:proposed}
\begin{figure*}
    \centering
    \includegraphics[width=\textwidth]{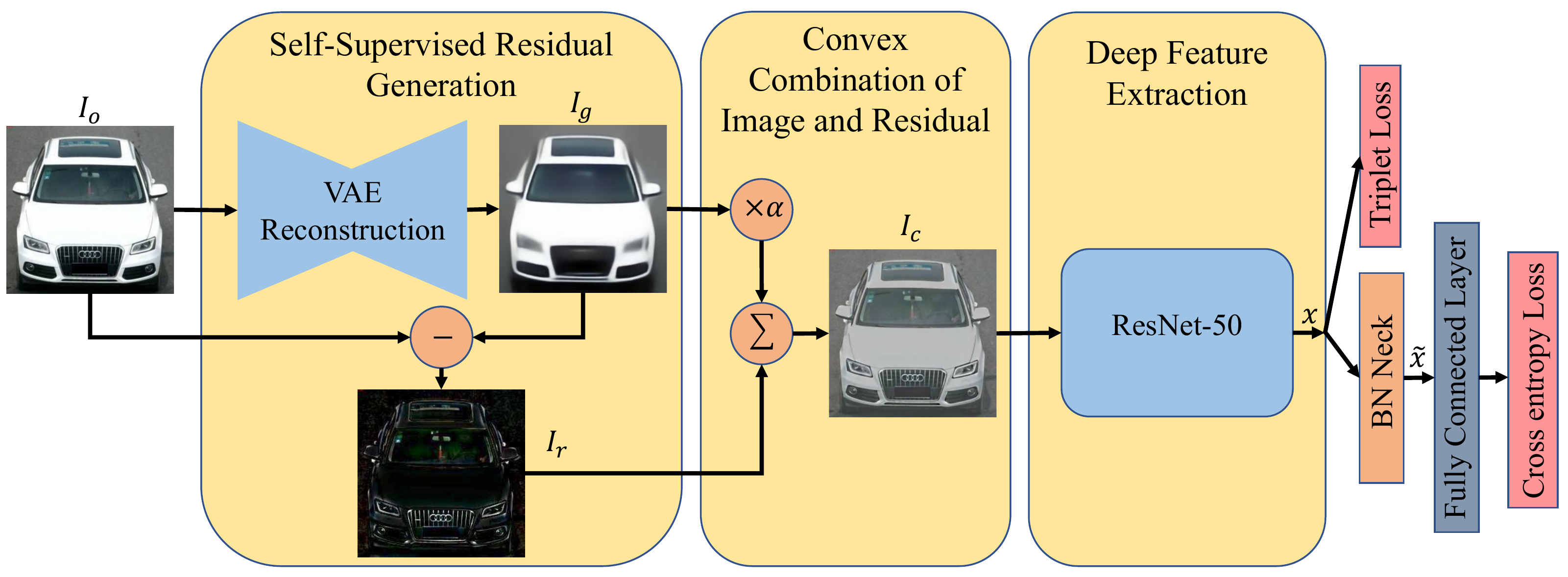}
    \caption{Proposed SAVER pipeline. The input image is passed through the VAE-based reconstruction module to remove vehicle-specific details. Next, the reconstruction is subtracted from the input image to form the residual image containing vehicle-specific details. Later, the convex combination (with trainable parameter $\alpha$) of the input and residual is calculated and passed through the re-id backbone for deep feature extraction. The entire pipeline is trained via triplet and cross entropy losses, separated via a batch normalization layer (BN Neck) proposed in \cite{luo2019bag}.}
    \label{fig:pipeline}
\end{figure*}

Our proposed pipeline is composed of two modules, namely, \textbf{Self-Supervised Residual Generation, and Deep Feature Extraction}. Fig. \ref{fig:pipeline} presents the proposed end-to-end pipeline. The self-supervised reconstruction network is responsible for creating the overall shape and structure of a vehicle image while obfuscating discriminative details. This enables us to highlight salient regions and remove background distractors by subtracting the reconstruction from the input image. Next, we feed the convex combination (with trainable parameter $\alpha$) of the residual and original input images to ResNet-50 \cite{kaiming2016resnet} model to generate robust discriminative features. To train our deep feature extraction module, we use techniques proposed in ``Bag of Tricks'' \cite{luo2019bag} and adapt them for vehicle re-identification, offering a strong baseline. 

\begin{figure*}
    \centering
    \includegraphics[width=\textwidth]{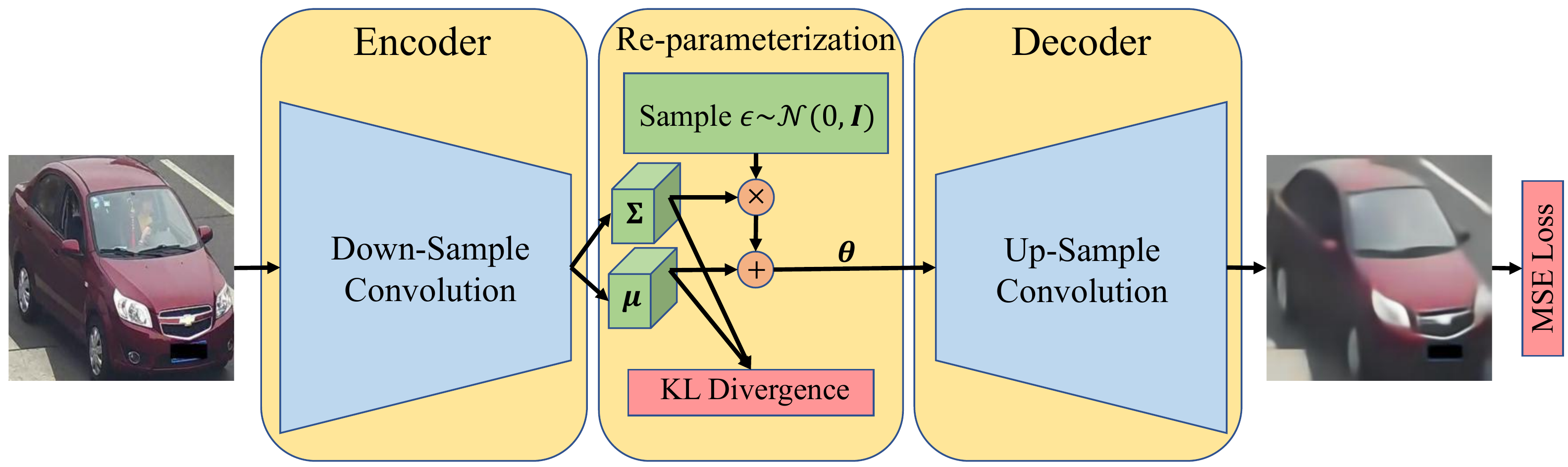}
    \caption{Self-Supervised image reconstruction required for subsequent residual generation. The input image goes through the convolutional encoder and is mapped to 3-dimensional latent variable. Using the VAE re-parameterization trick, a sample from the standard multivariate Gaussian $\epsilon$ is drawn and scaled via mean $\mu$ and co-variance $\mathbf{\Sigma}$ of the latent variable. Lastly, $\theta$ is up-sampled with a convolutional decoder to generate the input image template with most fine grained details removed.}
    \label{fig:reconstruction}
\end{figure*}

\subsection{Self-Supervised Residual Generation}
In order to generate the crude shape and structure of a vehicle while removing small-scale discriminative information, we leverage prior work in image segmentation \cite{BadrinarayananSegNet} and generation \cite{kingmaVAE}. Specifically, we construct a novel VAE architecture that down-samples the input image of spatial size $H \times W$ through max-pooling into a latent space of spatial size $\frac{H}{16} \times \frac{W}{16}$. Afterwards, we apply the re-parameterization trick introduced in \cite{kingmaVAE} to the latent features via their mean and covariance. Next, we up-sample the latent feature map as proposed by \cite{odena2016deconvolution} to prevent checkerboard artifacts. This step generates the reconstructed image of size $H \times W$. Fig. \ref{fig:reconstruction} illustrates the proposed self-supervised reconstruction network. 

Formally, we pre-train our reconstruction model using the mean squared error (MSE) and Kullback-Leibler (KL) divergence such that
\begin{equation}
\mathcal{L}_{reconstruction} = \mathcal{L}_{MSE} + \lambda\mathcal{L}_{KL}
\label{eq:reconstruction_loss}
\end{equation}
where 
\begin{equation}
     \mathcal{L}_{MSE} = \frac{1}{H \times W} \sum_{j=1}^{H} \sum_{k=1}^{W} |I_{o}(j, k) - I_{g}(j ,k)|^2
     \label{eq:MSE_loss}
\end{equation}
and 
\begin{equation}
    \mathcal{L}_{KL} = \frac{1}{2 \times (\frac{H}{16} \times \frac{W}{16})} \sum_{m=1}^{M}\left[ \mu_{m}^2 + \sigma_{m}^2 - \log(\sigma_{m}^2) -1 \right]
    \label{eq:KL}
\end{equation}
In Eq. \ref{eq:reconstruction_loss}, $\lambda$ sets the balance between the MSE and KL objective functions. Also, $I_o$ and $I_g$ in Eq. \ref{eq:MSE_loss} refer to the original and generated images respectively. Finally, in Eq. \ref{eq:KL}, $M$ is the dimensionality of the latent features $\mathbf{\theta} \in \mathbb{R}^{M}$ with mean $\mathbf{\mu} = [\mu_1, \dots, \mu_M]$ and covariance matrix $\mathbf{\Sigma} = \text{diag}({\sigma_1}^2,\dots,{\sigma_M}^2)$,
 that are re-parameterized via sampling from standard multivariate Gaussian $\epsilon \sim \mathcal{N}(\mathbf{0}, I_{M})$, \emph{i.e.} $\theta = \mathbf{\mu} + \mathbf{\Sigma}^{\frac{1}{2}} \mathbf{\epsilon}$.

We pre-train this model on the large-scale Vehicle Universe dataset, introduced in section \ref{subsubsec:universe}, prior to training our end-to-end pipeline, as described in section \ref{sec:experiments}. This pre-training allows the reconstruction model to generalize to vehicle images with a larger variety of make, model, color, orientation, and image quality. Hence, it captures domain invariant features that can later be fined-tuned for a particular dataset. Additionally, pre-training improves the rate of convergence for end-to-end pipeline training. It is important to note that unlike traditional VAE implementations, we use three-dimensional latent feature maps, \emph{i.e.}, channel, height and width dimensions, rather than one-dimensional latent vectors with only channel dimension, for improving the reconstruction quality and preserve more spatial information. Moreover, we scale $\mathcal{L}_{KL}$ when calculating Eq. \ref{eq:reconstruction_loss} to improve the reconstruction quality. We further explore the effect of the KL divergence scaling factor $\lambda$ in section \ref{sec:ablation}. Once the self-supervised image reconstruction network generates the coarse image template $I_g$, we subtract it from original input to obtain the residual image, \emph{i.e.} $I_r = I_o - I_g$.

\subsection{Deep Feature Extraction}
\label{subsec:deep_feature_extraction}
Since vehicle images reside on a high-dimensional manifold, we employ a DCNN to project the images onto a lower-dimensional vector space while preserving features that can effectively characterize a unique vehicle identity. To this end, we use a single-branch ResNet-50. To train this model, we use techniques proposed in ``Bag of Tricks'' \cite{luo2019bag}, which are shown to help a DCNN traverse the optimization landscape using gradient-based optimization methods more effectively. In particular, we observe that the following techniques significantly contribute to the performance of the vehicle re-id baseline model: 
\begin{itemize}
    \item [1 -] \textbf{Learning Rate Warm-Up}: \cite{fan2019spherereid} has suggested increasing the learning rate linearly in initial epochs of training to obtain improved weight initialization. This significantly contributes to the enhanced performance of our baseline.
    \item [2 -] \textbf{Random Erasing Augmentation (REA)}: To better handle the issue of occlusion, \cite{hermans2017defense} introduced REA with the goal of encouraging a network to learn more robust representations.
    \item [3 -] \textbf{Label Smoothing}: In order to alleviate the issue of over-fitting to the training data, \cite{szegedy2016rethinking} proposed smoothing the ground-truth labels. 
    \item [4 -] \textbf{Batch Normalization (BN) Neck}: To effectively apply both classification and triplet losses to the extracted features, a BN layer is proposed by \cite{luo2019bag}. This also significantly improves vehicle re-id performance.
\end{itemize}

The ResNet-50 feature extractor model is trained to optimize for triplet and cross entropy classification losses which are calculated as follows:
\begin{equation}
    \label{eq:triplet}
    \centering
    \mathcal{L}_{triplet} = \frac{1}{B} \sum_{i=1}^{B} \sum_{a \in b_{i}} \left[\gamma + \max_{p \in \mathcal{P}(a)} d(x_a,x_p) - \min_{n \in \mathcal{N}(a)} d(x_a,x_n)  \right]_{+}
\end{equation}
and
\begin{equation}
    \label{eq:cls}
    \centering
    \mathcal{L}_{classification} = - \frac{1}{N} \sum_{i=1}^{N} \log \frac{\exp^(W_{c(\hat{x}_i)}^T \Tilde{x}_i + b_{c(\hat{x}_i)})}{\sum_{j = 1}^{C} \exp^(W_j^T \Tilde{x}_i + b_j)}
\end{equation}
In Eq. \ref{eq:triplet}, $B$, $b_i$, $a$, $\gamma$, $\mathcal{P}(a)$ and $\mathcal{N}(a)$ are the total number of batches, $i^{th}$ batch, anchor sample, distance margin threshold, positive and negative sample sets corresponding to a given anchor respectively. Moreover, $x_a, x_p, x_n$ represent the ResNet-50 extracted features associated with anchor, positive and negative samples. In addition, function $d(.,.)$ calculates the Euclidean distance of the two extracted features. Note that in Eq. \ref{eq:triplet}, we used the batch hard triplet loss~\cite{hermans2017defense} to overcome the computational complexity of calculating the distances to all unique triplets of data points. Here we construct batches so that they have exactly $K$ instances of each ID used in a particular batch, \emph{i.e.} $B$ is a multiple of $K$.
In Eq. \ref{eq:cls}, $\Tilde{x}_i$ and $c(\hat{x}_i)$ refer to the extracted feature for the $i^{th}$ image in the training set after passing through the BN Neck layer and its corresponding ground-truth class label respectively. Furthermore, $W_j$, $b_j$ are the weight vector and bias associated with class $j$ in the final classification layer. $N$ and $C$ represent the total number of samples and classes in the training process respectively.

\subsection{End-To-End Training}
After pre-training the self-supervised residual generation module, we jointly train the VAE and deep feature extractor. We compute the convex combination of input images and their respective residuals using a learnable parameter $\alpha$, \emph{i.e.} $I_c = \alpha \times I_o + (1- \alpha) \times I_r$, allowing the feature extractor network to weight the importance of each input source. Moreover, the end-to-end training helps the entire pipeline adapt the residual generation such that it is suited for the re-id task. In summary, the loss function for end-to-end training is the following:
\begin{equation}
    \mathcal{L}_{total} = \mathcal{L}_{triplet} + \mathcal{L}_{classification} + \eta \mathcal{L}_{reconstruction}
    \label{eq:loss_total}
\end{equation}
In Eq. \ref{eq:loss_total}, the scaling factor $\eta$ is empirically set to $100$. 

\section{Experiments}
\label{sec:experiments}
In this section, we first present the different datasets on which we evaluate the proposed approach and describe how vehicle re-identification systems are evaluated in general. Next, we present implementation details for the  proposed self-supervised residual generation, deep feature extraction and end-to-end training steps. Finally, we report experimental results of the proposed approach.

\subsection{Vehicle Re-Identification Datasets}
\label{subsec:dataset}

We evaluate SAVER on six popular vehicle re-id benchmarks, including VeRi, VehicleID, VERI-Wild, Vehicle-1M and PKU VD1$\&$VD2. Table \ref{tab:datasets} presents the statistics of these datasets in terms of the number of unique identities, images and cameras. Additionally, we highlight four additional datasets of unconstrained vehicle images, including CityFlow, CompCars, BoxCars116K \cite{Sochor2018_BoxCars}, and StanfordCars \cite{krause2013_StanfordCars}, used in the pre-training of our self-supervised reconstruction network.
\begin{table*}
    \centering
    \caption{Vehicle re-id datasets statistics. ID, IM, Cam refer to number of unique identities, images and cameras respectively. Note that the evaluation set of VehicleID, VERI-Wild, Vehicle-1M, VD1 $\&$ VD2 are partitioned into small (S), medium (M) and large (L) splits.}
    \resizebox{\columnwidth}{!}{%
    \begin{tabular}{c|c|c|c|c|c|c|c|c|c|c|c|c|c|c|c|c|c|}
         \cline{3-18}
         \multicolumn{2}{}{} & \multicolumn{16}{|c|}{Vehicle Re-id Datasets} \\
         \cline{1-18}
         \multicolumn{2}{|c|}{Split Set} & \multicolumn{1}{|c|}{VeRi} & \multicolumn{3}{|c|}{VehicleID} & \multicolumn{3}{|c|}{VERI-Wild} & \multicolumn{3}{|c|}{Vehicle-1M} & \multicolumn{3}{|c|}{VD1} & \multicolumn{3}{|c|}{VD2} \\
         \cline{1-18} \cline{1-18}
         \multicolumn{1}{|c|}{\multirow{3}{*}{\begin{turn}{90}Train\end{turn}}} & \multicolumn{1}{|c|}{ID} & \multicolumn{1}{|c|}{576} & \multicolumn{3}{|c|}{13164} & \multicolumn{3}{|c|}{30671} & \multicolumn{3}{|c|}{50000} & \multicolumn{3}{|c|}{70591} & \multicolumn{3}{|c|}{39619} \\
         \cline{2-18}
         \multicolumn{1}{|c|}{} & \multicolumn{1}{|c|}{IM} &\multicolumn{1}{|c|}{37746} & \multicolumn{3}{|c|}{113346} & 
 \multicolumn{3}{|c|}{277797} & \multicolumn{3}{|c|}{844571} & \multicolumn{3}{|c|}{422326} & \multicolumn{3}{|c|}{342608} \\
         \cline{2-18}
         \multicolumn{1}{|c|}{} & \multicolumn{1}{|c|}{Cam} &\multicolumn{1}{|c|}{20} & \multicolumn{3}{|c|}{-} & \multicolumn{3}{|c|}{173} & \multicolumn{3}{|c|}{-} & \multicolumn{3}{|c|}{-} & \multicolumn{3}{|c|}{-} \\
         \cline{1-18} \cline{1-18}
         \multicolumn{1}{|c|}{\multirow{4}{*}{\begin{turn}{90}Gallery\end{turn}}} & 
         \multicolumn{1}{|c|}{} & \multicolumn{1}{|c|}{} & \multicolumn{1}{|c|}{S} &  \multicolumn{1}{|c|}{M} & \multicolumn{1}{|c|}{L} & \multicolumn{1}{|c|}{S} &  \multicolumn{1}{|c|}{M} & \multicolumn{1}{|c|}{L} & \multicolumn{1}{|c|}{S} &  \multicolumn{1}{|c|}{M} & \multicolumn{1}{|c|}{L} & \multicolumn{1}{|c|}{S} &  \multicolumn{1}{|c|}{M} & \multicolumn{1}{|c|}{L} & \multicolumn{1}{|c|}{S} &  \multicolumn{1}{|c|}{M} & \multicolumn{1}{|c|}{L} \\
         \cline{4-18} \cline{4-18} \cline{4-18}
         \multicolumn{1}{|c|}{} & \multicolumn{1}{|c|}{ID} & \multicolumn{1}{|c|}{200} & \multicolumn{1}{|c|}{800} & \multicolumn{1}{|c|}{1600} & \multicolumn{1}{|c|}{2400} & \multicolumn{1}{|c|}{3000} & \multicolumn{1}{|c|}{5000} & \multicolumn{1}{|c|}{10000} & \multicolumn{1}{|c|}{1000} & \multicolumn{1}{|c|}{2000} & \multicolumn{1}{|c|}{3000} & \multicolumn{1}{|c|}{18000} & \multicolumn{1}{|c|}{131275} & \multicolumn{1}{|c|}{141757} & \multicolumn{1}{|c|}{12000} & \multicolumn{1}{|c|}{70755} & \multicolumn{1}{|c|}{79764} \\
         \cline{2-18}
         \multicolumn{1}{|c|}{} & \multicolumn{1}{|c|}{IM} & \multicolumn{1}{|c|}{11579} & \multicolumn{1}{|c|}{800} & \multicolumn{1}{|c|}{1600} & \multicolumn{1}{|c|}{2400} & \multicolumn{1}{|c|}{38861} & \multicolumn{1}{|c|}{64389} & \multicolumn{1}{|c|}{128517} & \multicolumn{1}{|c|}{1000} & \multicolumn{1}{|c|}{2000} & \multicolumn{1}{|c|}{3000} & \multicolumn{1}{|c|}{104887} & \multicolumn{1}{|c|}{602032} & \multicolumn{1}{|c|}{1095649} & \multicolumn{1}{|c|}{103550} & \multicolumn{1}{|c|}{455910} & \multicolumn{1}{|c|}{805260} \\
         \cline{2-18}
         \multicolumn{1}{|c|}{} & \multicolumn{1}{|c|}{Cam} & \multicolumn{1}{|c|}{19} & \multicolumn{1}{|c|}{-} & \multicolumn{1}{|c|}{-} & \multicolumn{1}{|c|}{-} & \multicolumn{1}{|c|}{146} & \multicolumn{1}{|c|}{153} & \multicolumn{1}{|c|}{161} & \multicolumn{1}{|c|}{-} & \multicolumn{1}{|c|}{-} & \multicolumn{1}{|c|}{-} & \multicolumn{1}{|c|}{-} & \multicolumn{1}{|c|}{-} & \multicolumn{1}{|c|}{-} & \multicolumn{1}{|c|}{-} & \multicolumn{1}{|c|}{-} & \multicolumn{1}{|c|}{-} \\
         \cline{1-18}
         \multicolumn{1}{|c|}{\multirow{2}{*}{\begin{turn}{90}Query\end{turn}}} & \multicolumn{1}{|c|}{ID} & \multicolumn{1}{|c|}{200} & \multicolumn{1}{|c|}{800} & \multicolumn{1}{|c|}{1600} & \multicolumn{1}{|c|}{2400} & \multicolumn{1}{|c|}{3000} & \multicolumn{1}{|c|}{5000} & \multicolumn{1}{|c|}{10000} & \multicolumn{1}{|c|}{1000} & \multicolumn{1}{|c|}{2000} & \multicolumn{1}{|c|}{3000} & \multicolumn{1}{|c|}{2000} & \multicolumn{1}{|c|}{2000} & \multicolumn{1}{|c|}{2000} & \multicolumn{1}{|c|}{2000} & \multicolumn{1}{|c|}{2000} & \multicolumn{1}{|c|}{2000} \\
         \cline{2-18}
         \multicolumn{1}{|c|}{} & \multicolumn{1}{|c|}{IM} & \multicolumn{1}{|c|}{1678} & \multicolumn{1}{|c|}{5693} & \multicolumn{1}{|c|}{11777} & \multicolumn{1}{|c|}{17377} & \multicolumn{1}{|c|}{3000} & \multicolumn{1}{|c|}{5000} & \multicolumn{1}{|c|}{10000} & \multicolumn{1}{|c|}{15123} & \multicolumn{1}{|c|}{30539} & \multicolumn{1}{|c|}{45069} & \multicolumn{1}{|c|}{2000} & \multicolumn{1}{|c|}{2000} & \multicolumn{1}{|c|}{2000} & \multicolumn{1}{|c|}{2000} & \multicolumn{1}{|c|}{2000} & \multicolumn{1}{|c|}{2000} \\
         \cline{2-18}
         \multicolumn{1}{|c|}{} & \multicolumn{1}{|c|}{Cam} & \multicolumn{1}{|c|}{19} & \multicolumn{1}{|c|}{-} & \multicolumn{1}{|c|}{-} & \multicolumn{1}{|c|}{-} & \multicolumn{1}{|c|}{105} & \multicolumn{1}{|c|}{113} & \multicolumn{1}{|c|}{126} & \multicolumn{1}{|c|}{-} & \multicolumn{1}{|c|}{-} & \multicolumn{1}{|c|}{-} & \multicolumn{1}{|c|}{-} & \multicolumn{1}{|c|}{-} & \multicolumn{1}{|c|}{-} & \multicolumn{1}{|c|}{-} & \multicolumn{1}{|c|}{-} & \multicolumn{1}{|c|}{-} \\
         \cline{1-18}
    \end{tabular}
    }
    \label{tab:datasets}
\end{table*}

Re-id systems are commonly evaluated using the Cumulative Match Curve (CMC) and Mean Average Precision (mAP). A fixed gallery set is ranked with respect to the similarity score, e.g., $L_2$ distance, of its images and a given query image. CMC@K measures the probability of having a vehicle with the same ID as the query within the top K elements of the ranked gallery. It is a common practice to report CMC@$1$ and CMC@$5$. 
Similarly, mAP measures the average precision over all images in a query set. 

\subsection{Implementation Details}
Here we discuss the implementation of both the self-supervised residual generation and deep feature extraction modules. In general, we resize all the images to $(256,256)$ and normalize them by a mean and standard deviation of $0.5$ across RGB channels before passing them through the respective networks. Moreover, similar to \cite{Khorramshahi_2019_CVPR_Workshops}, we pre-process all images across all the experiments with the Detectron object detector \cite{Detectron2018} to minimize background noise. 

\subsubsection{Self-Supervised Residual Generation}
\label{subsubsec:universe}
To pre-train the self-supervised residual generation module, we construct the large-scale Vehicle Universe dataset. We specifically consider vehicles from a variety of distributions to improve the robustness of our model. We utilize data from several sources, including CompCars, StanfordCars, BoxCars116K, CityFlow, PKU VD1$\&$VD2, Vehicle-1M, VehicleID, VeRi and VeRi-Wild. In total, Vehicle Universe has $3706670$, $1103404$ and $11146$ images in the train, test and validation sets respectively.

\subsubsection{Deep Feature Extraction}
As mentioned in section \ref{subsec:deep_feature_extraction}, we use ResNet-50 for feature extraction. 
In all of our experiments learning rate starts from $3.5e-5$ and is linearly increased with the slope of $3.1e-5$ 
in the first ten epochs. Afterwards, it is decayed by a factor of ten every $30^{th}$ epoch. In total, the end-to-end pipeline is trained for $150$ epochs via Adam \cite{kingma2014adam} optimizer. Furthermore, we use an initial value of $\alpha=0.5$ for convex combination and $\gamma=0.3$ for the triplet loss in Eq. \ref{eq:triplet}.

\subsection{Experimental Evaluation}
In this section, we present evaluation results of the global appearance model (baseline) and global appearance model augmented with self-supervised attention (SAVER) on different re-id benchmarks discussed in section \ref{subsec:dataset}.
\subsubsection{VeRi}
Table \ref{tab:veri_results} reports the evaluation results on VeRi, a popular dataset for vehicle re-id. SAVER improves upon the strong baseline model. Most notably, SAVER gives $1.4\%$ improvement on the mAP metric. We note that $\alpha$ in the convex combination of the input and residual saturates at $0.96$, which means the model relies on $96\%$ percent of the original image and $4\%$ of the residual to construct more robust features.

\subsubsection{VehicleID}
\begin{figure}[t]
    \centering
    \subfloat[]{\includegraphics[width=0.15\textwidth, height=0.15\textwidth]{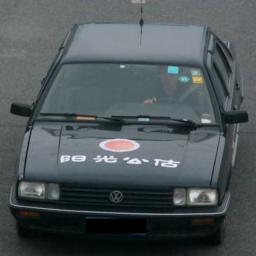}}~
    \subfloat[]{\includegraphics[width=0.15\textwidth, height=0.15\textwidth]{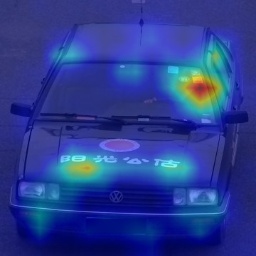}}~
    \subfloat[]{\includegraphics[width=0.15\textwidth, height=0.15\textwidth]{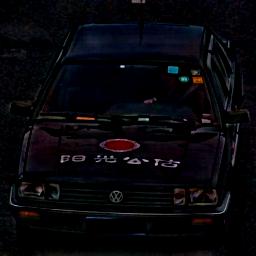}}~
    \subfloat[]{\includegraphics[width=0.15\textwidth, height=0.15\textwidth]{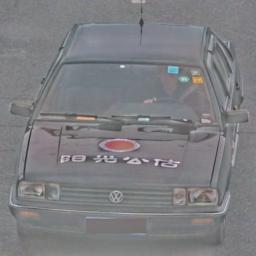}}~
    \subfloat[]{\includegraphics[width=0.15\textwidth, height=0.15\textwidth]{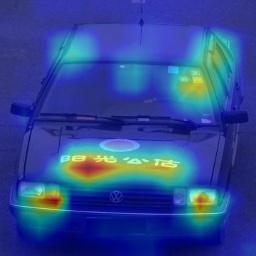}}
    \caption{Grad-CAM visualization of baseline and SAVER; (a) original image, (b) Grad-CAM visualization corresponding to the baseline model, \emph{e.i.}, $\alpha=1$, (c),(d) are residual and normalized residual maps (for the sake of visualization) obtain via our proposed self-supervised model respectively. (e) is the Grad-CAM visualization of proposed model, \emph{e.i.}, $\alpha=0.97$ in VehicleID dataset.}
    \label{fig:VehicleID_gradcam}
\end{figure}

Table \ref{tab:vehicleid_results} presents results of baseline and SAVER on test sets of varying sizes. Performance improvement of $+1.0\%$ in CMC@1 over the baseline model can be observed for all the test splits. 
To better demonstrate the discriminating capability of the proposed model, we visualize the attention map of both baseline and the proposed SAVER models on an image of this dataset using Gradient Class Activation Mapping (Grad-CAM) \cite{selvaraju2017grad}. In Figure \ref{fig:VehicleID_gradcam}, it is clear that SAVER is able to effectively construct attention on regions containing discriminative information such as headlights, hood and windshield stickers.

\subsubsection{VERI-Wild}
Evaluation results on the VERI-Wild dataset are presented in Table \ref{tab:veriwild_results}. Notably, our proposed residual generation model is improved upon the baseline by $+2.0\%$ and $+1.0\%$ for mAP and CMC@1 metrics on all evaluation splits respectively. The final alpha value $\alpha=0.94$ suggests that the residual information contributes more in extracting robust features in this dataset.

\subsubsection{Vehicle-1M}
Table \ref{tab:vehicle1m_results} reports the results of baseline and the proposed methods. Similar to VehicleID dataset, Vehicle-1M does not include fixed evaluation sets, therefore we randomly construct the evaluation splits and keep them fixed throughout the experiments. With the value of $\alpha=0.98$ the proposed self-supervised residual generation module improves upon the baseline model in all metrics across all evaluation sets. 

\subsubsection{PKU VD1$\&2$}
\begin{figure}
    \centering
    \subfloat[][Query]{\includegraphics[width=0.15\textwidth, height=0.14\textwidth]{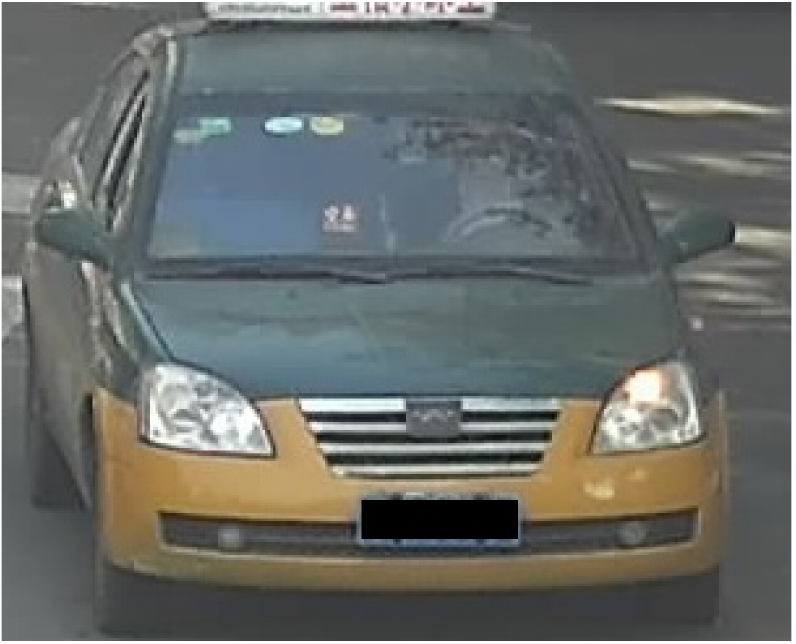}}\quad
    \subfloat[][Top 1]{\includegraphics[width=0.15\textwidth, height=0.14\textwidth]{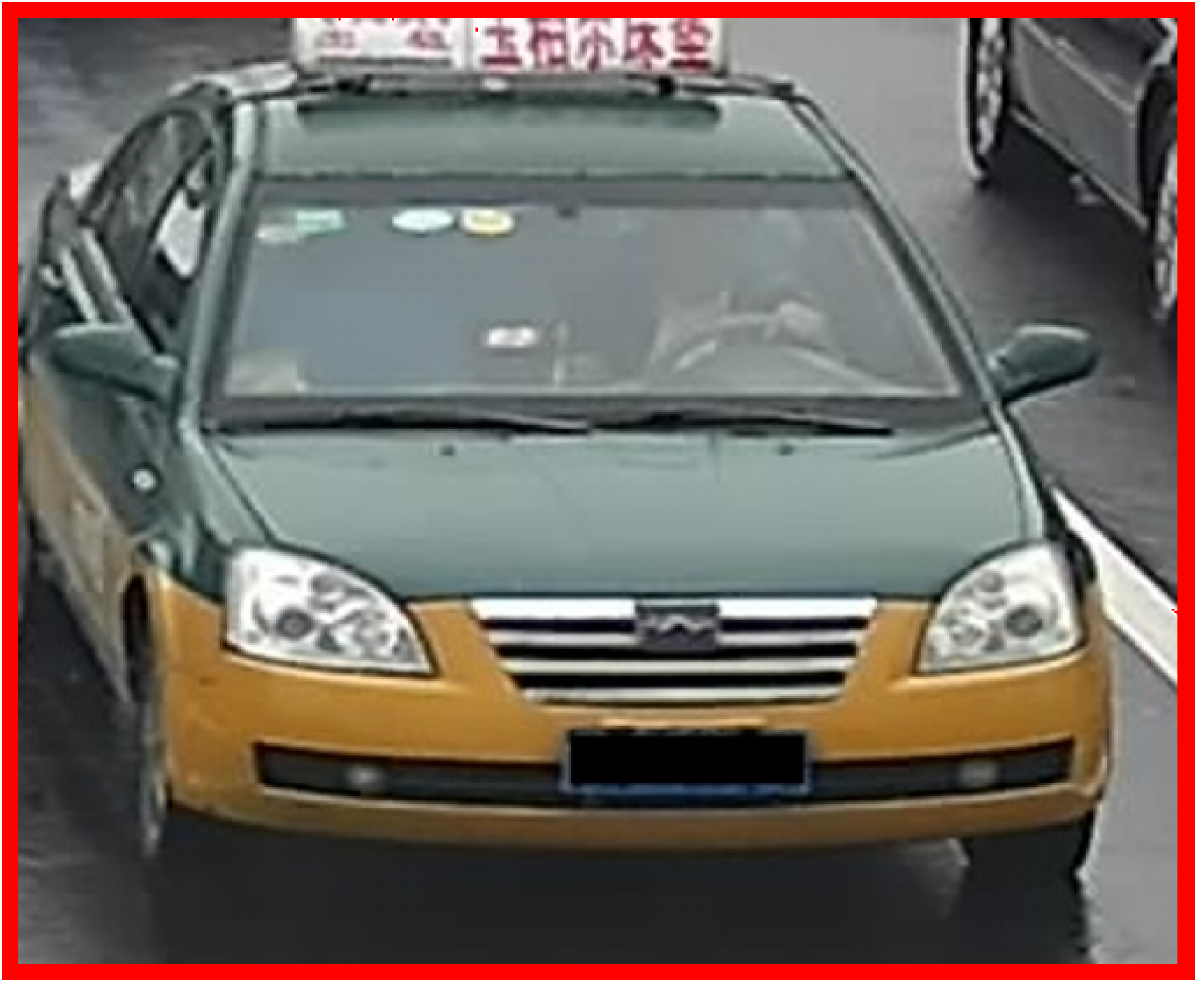}}\quad
    \subfloat[][Top 2]{\includegraphics[width=0.15\textwidth, height=0.14\textwidth]{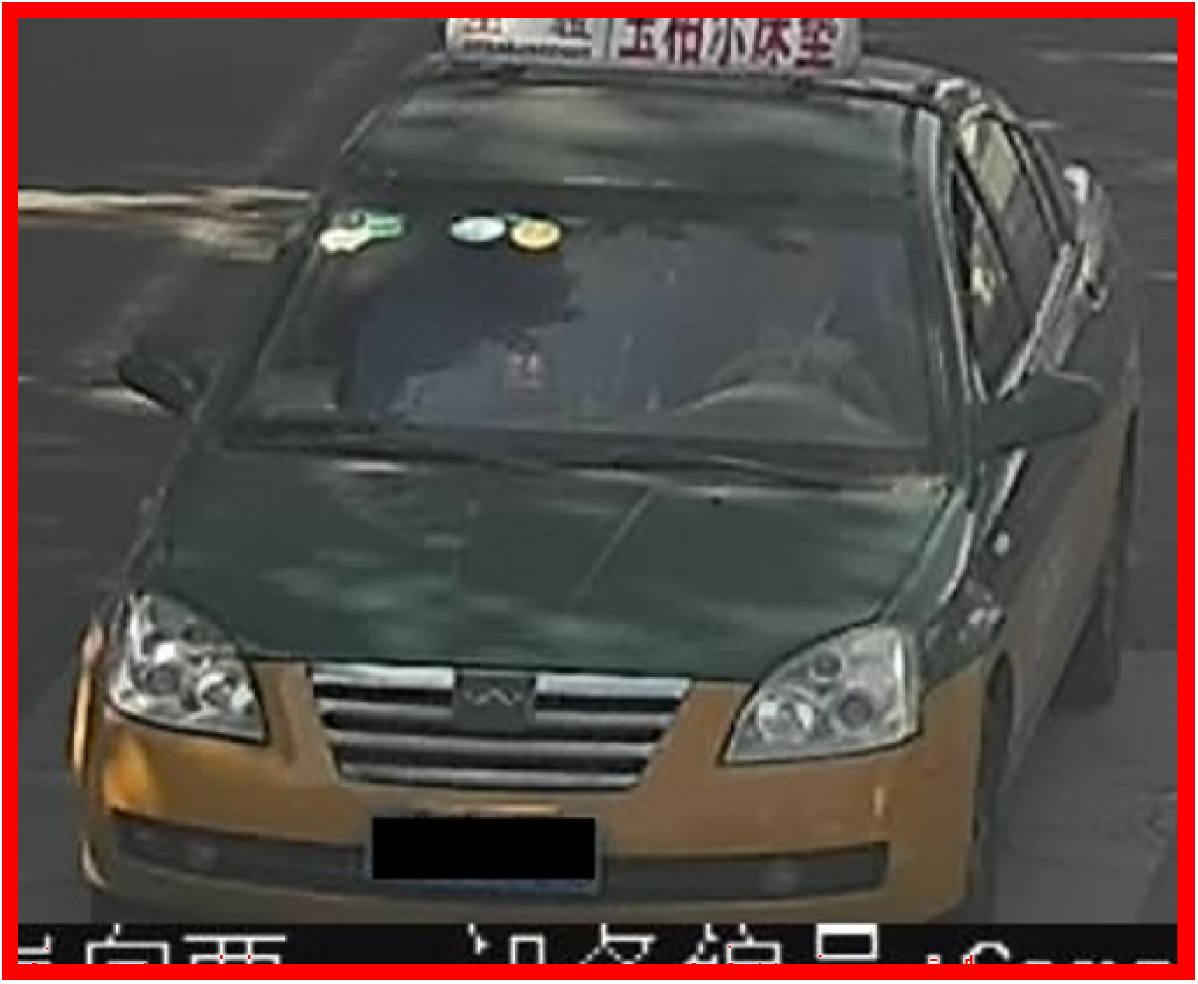}}\quad
    \subfloat[][Top 3]{\includegraphics[width=0.15\textwidth, height=0.14\textwidth]{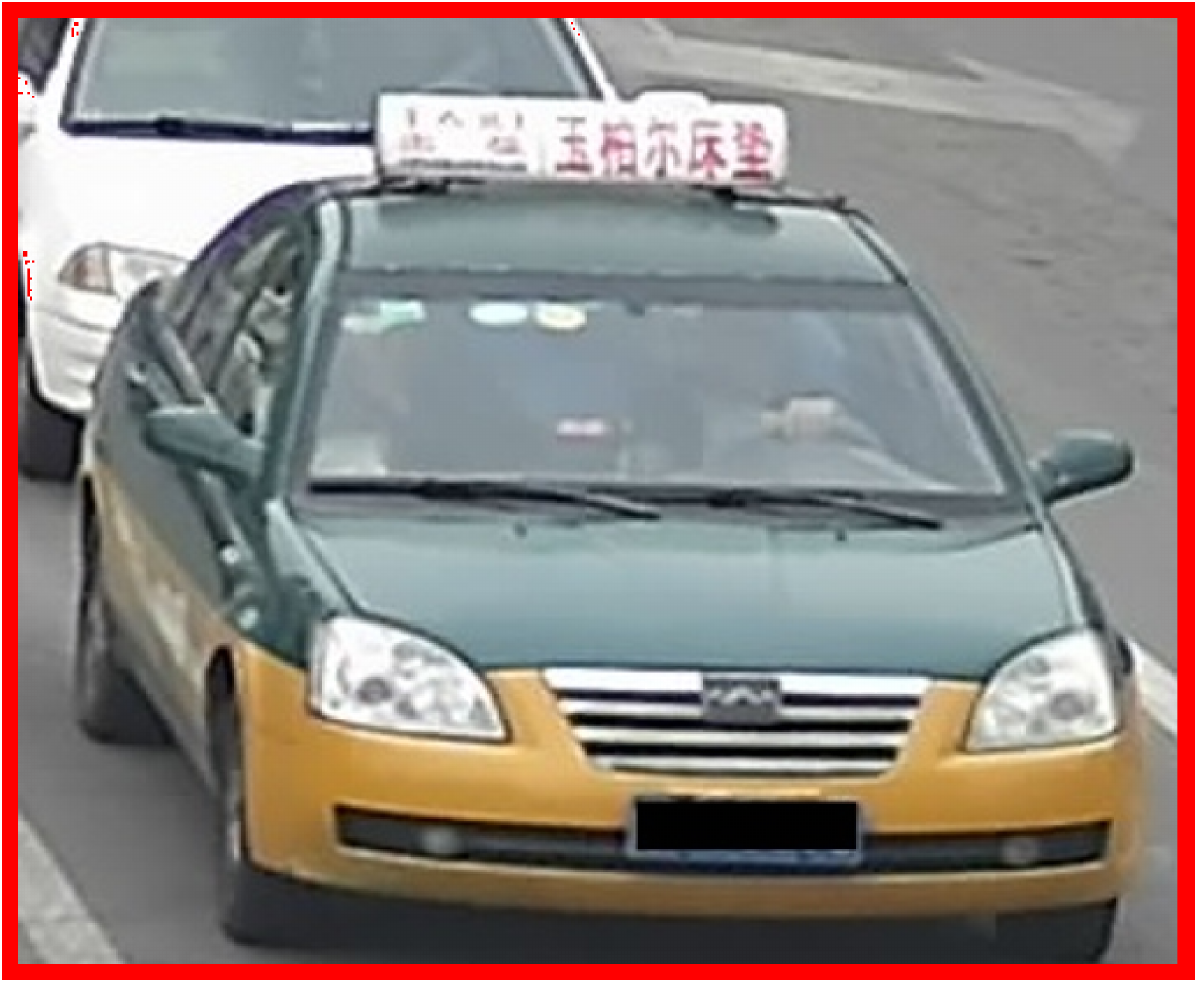}}
    \\
    \subfloat[][Query]{\includegraphics[width=0.15\textwidth, height=0.14\textwidth]{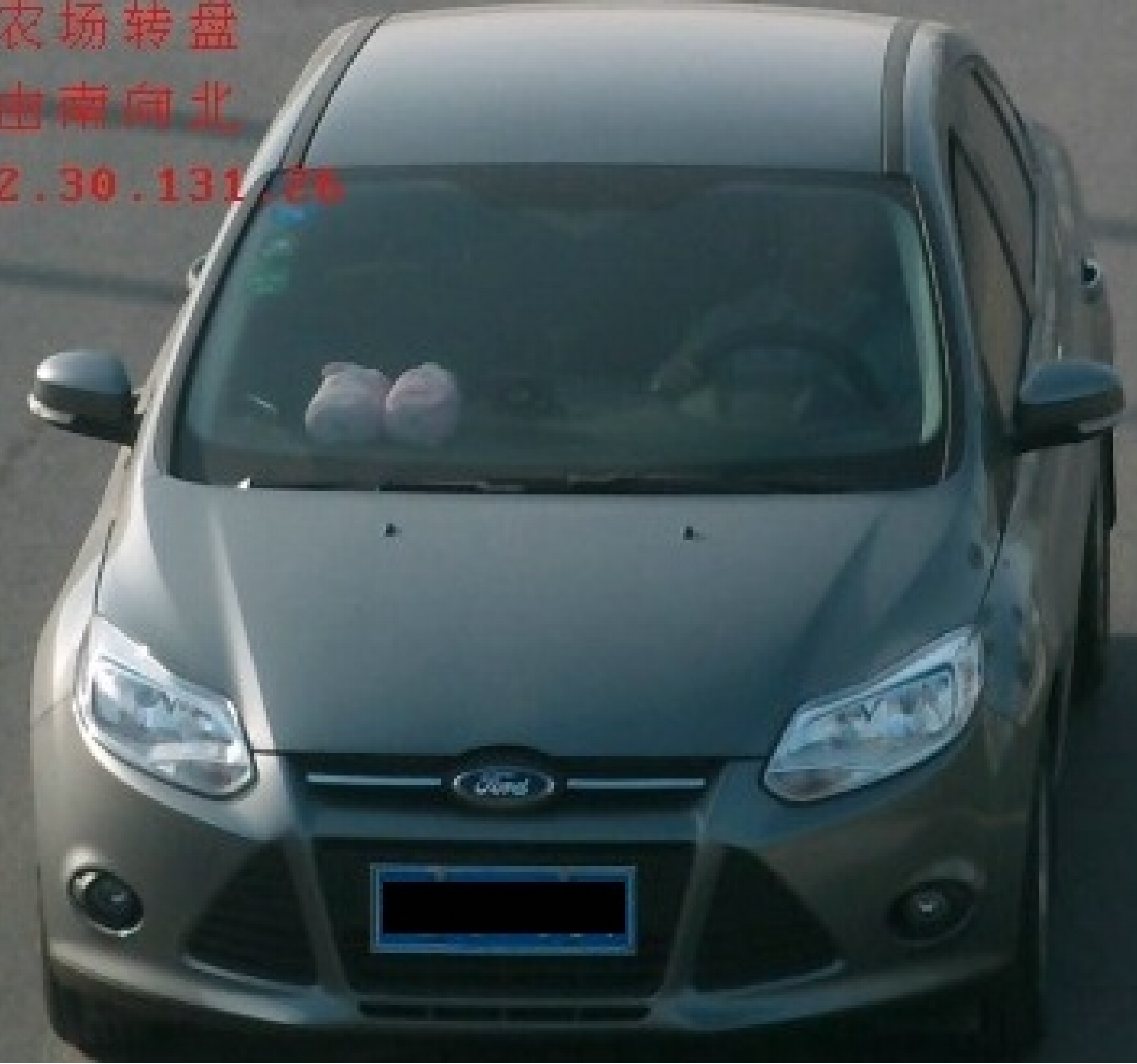}}\quad
    \subfloat[][Top 1]{\includegraphics[width=0.15\textwidth, height=0.14\textwidth]{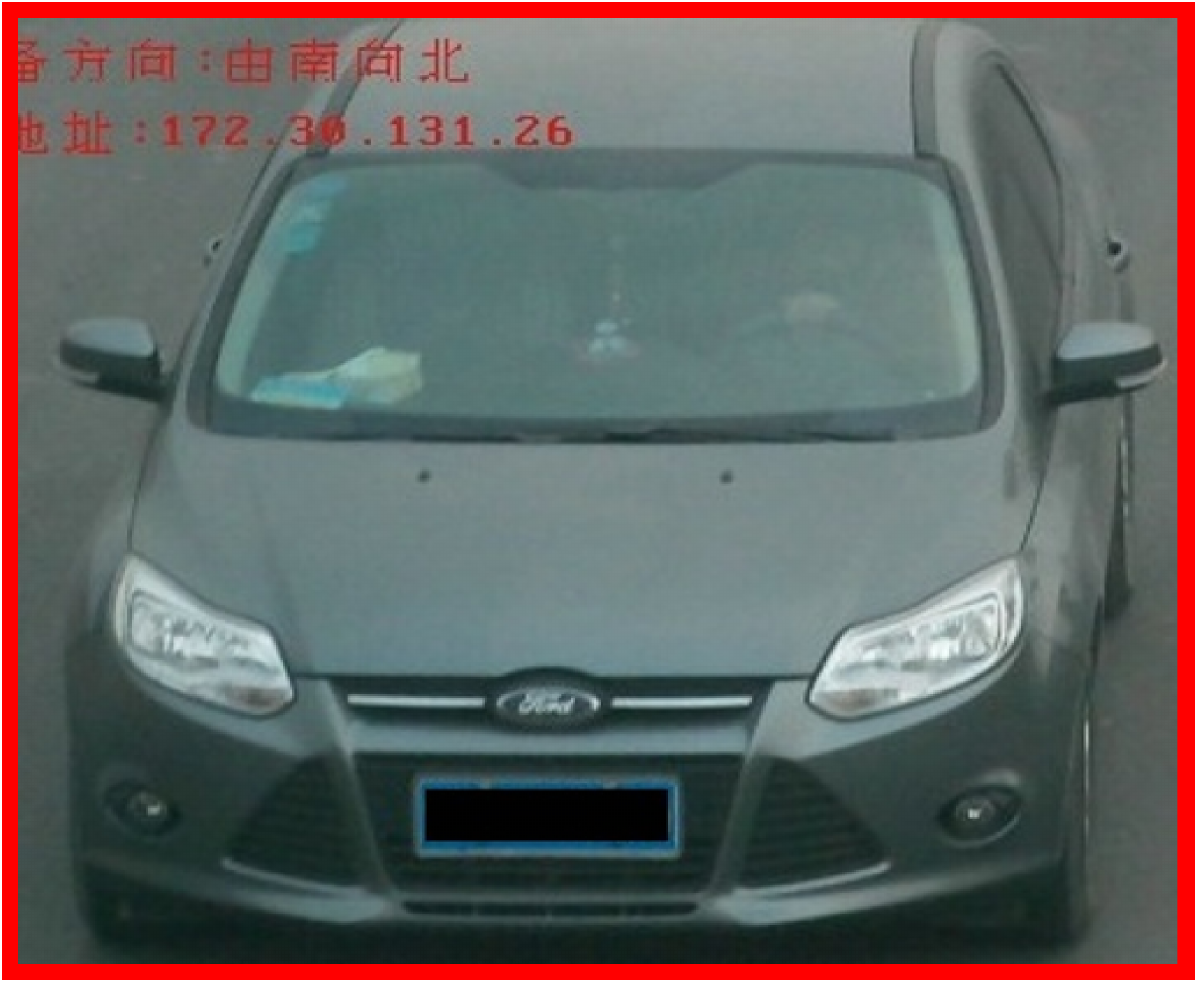}}\quad
    \subfloat[][Top 2]{\includegraphics[width=0.15\textwidth, height=0.14\textwidth]{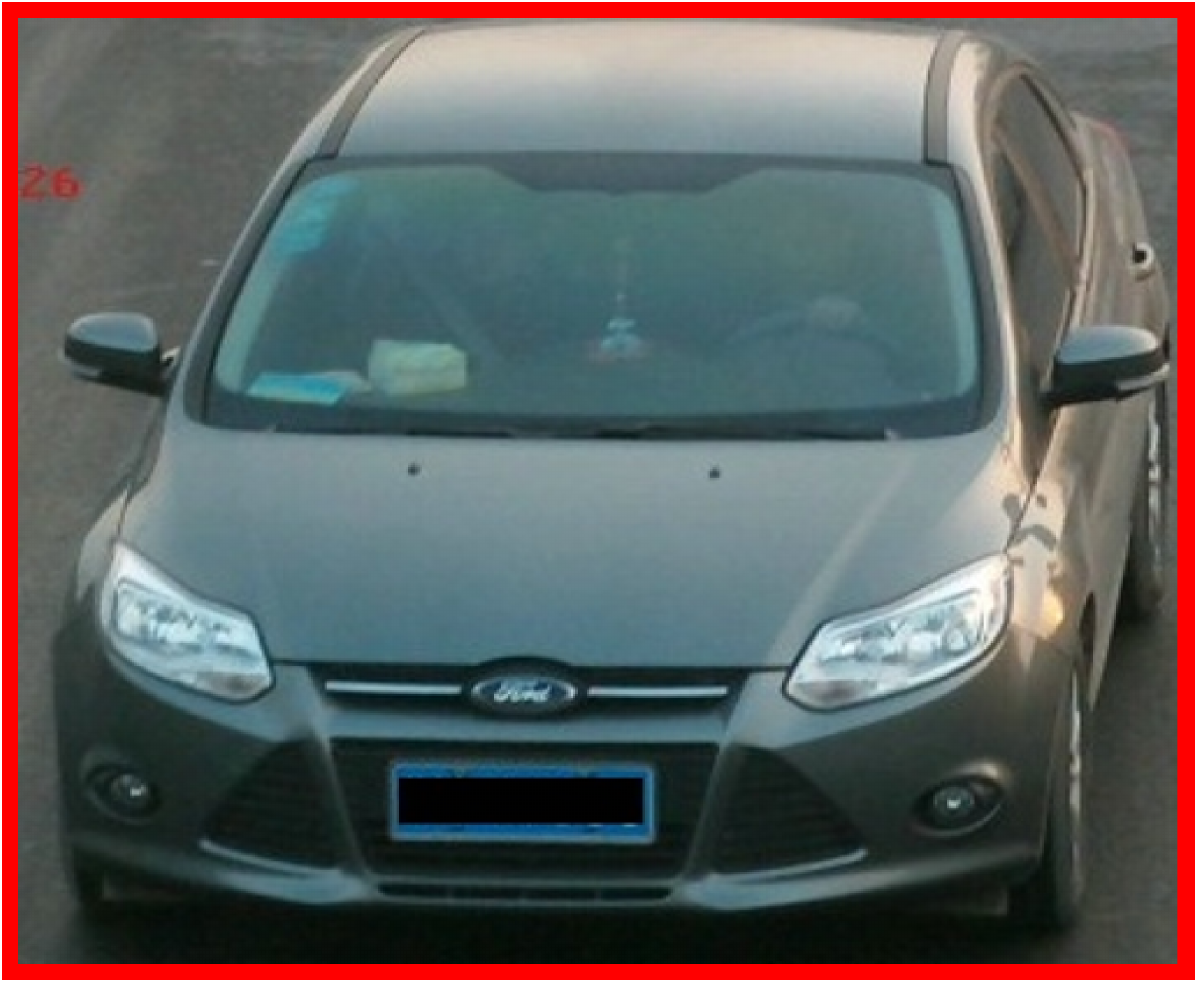}}\quad
    \subfloat[][Top 3]{\includegraphics[width=0.15\textwidth, height=0.14\textwidth]{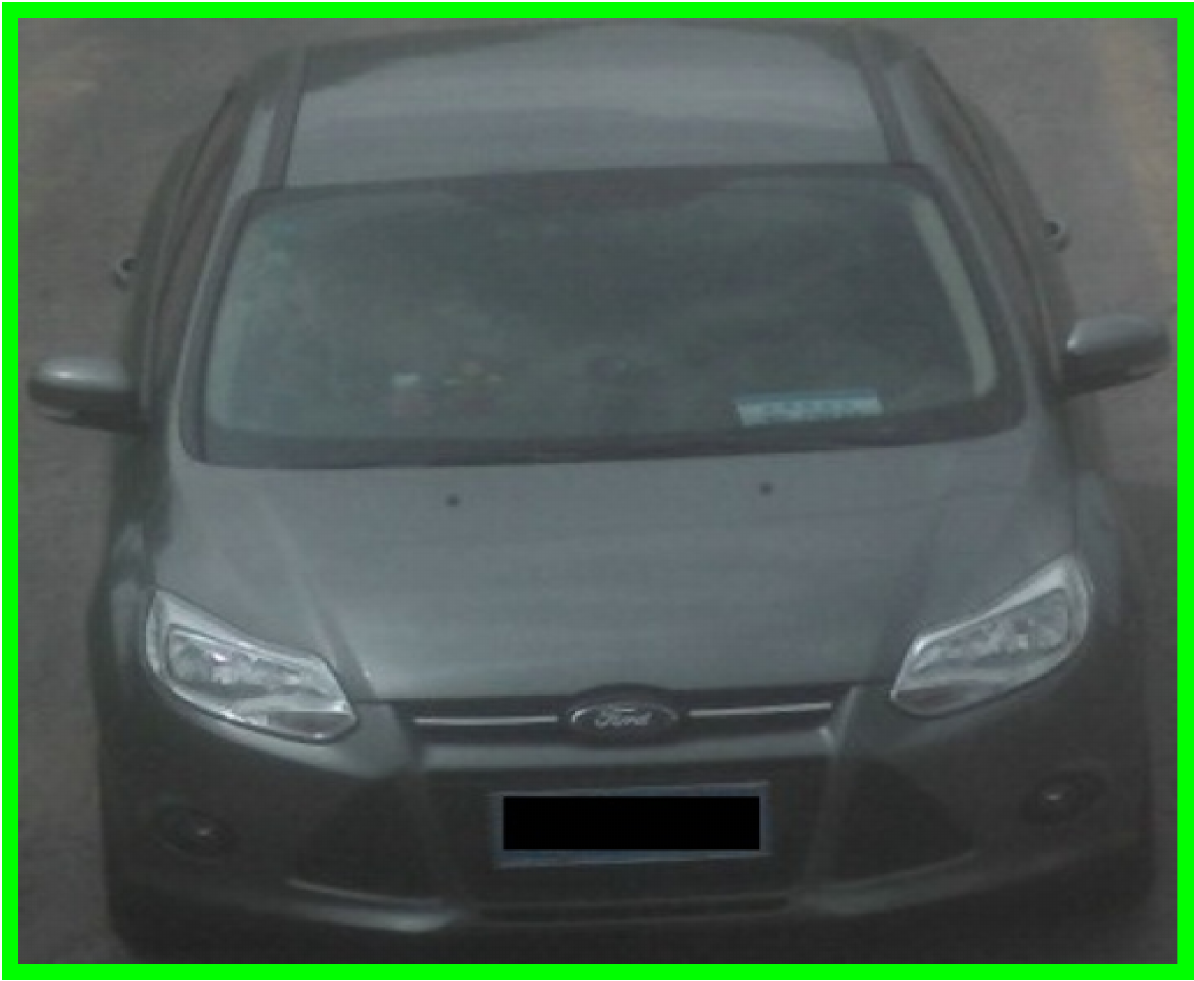}}
    \caption{Examples of SAVER failure on VD1 (sub-figures (a-d)) and VD2 (sub-figures (e-h)). The overall appearance of the query and top ranked images of the gallery are nearly identical. Visual cues such as windshield sticker placement are almost indistinguishable.}
    \label{fig:VD_failure}
\end{figure}

Table \ref{tab:VD} highlights the evaluation results on both PKU VD datasets. Similar to most re-id datasets, VD1$\&$2 have S/M/L evaluation sets. However, due to the extreme size of these data splits, as shown in Table \ref{tab:datasets}, we are only able to report numbers on the small evaluation set. The performance of SAVER is comparable to our baseline model. Moreover, the final value of $\alpha=0.99$ indicates that baseline models is already very strong, and has almost no room for improvement. We can conclude that our performance on these data sets are saturated. Qualitatively, in Figure \ref{fig:VD_failure} we show two failure cases of SAVER on these datasets. Note that how extremely similar these images are and it is nearly impossible to differentiate them based on only visual information. 
\begin{table*}
    \centering
    \caption{Performance Comparison on VeRi}
    \resizebox{0.65\columnwidth}{!}{%
    \begin{tabular}{|c|c|c|c|c}
    \cline{1-4}
    \multicolumn{1}{|c||}{Model} & mAP(\%) & CMC@1(\%) & CMC@5(\%) &  \\
    \cline{1-4}
    \multicolumn{1}{|c||}{Baseline}  & 78.2  & 95.5  & 97.9 &  \\
    \cline{1-5}
    \hline
    \multicolumn{1}{|c||}{SAVER} & \textbf{79.6}  & \textbf{96.4}  & \textbf{98.6} & \multicolumn{1}{c|}{$\alpha=0.96$} \\
    \hline
    \end{tabular}
    }
    \label{tab:veri_results}
\end{table*}
\begin{table*}
    \centering
    \caption{Performance Comparison on VehicleID}
    \resizebox{0.65\columnwidth}{!}{%
    \begin{tabular}{c|c|c|c|c|c|c|c}
    \cline{1-7}
    \multicolumn{1}{|c||}{\multirow{2}{*}{Model}} &  \multicolumn{3}{|c|}{CMC@1(\%)} & \multicolumn{3}{|c|}{CMC@5(\%)} & \\
    \cline{2-7}
    \multicolumn{1}{|c||}{} & S & M & L & S & M & L &\\
    \cline{1-7}
    \multicolumn{1}{|c||}{Baseline} & 78.4 & 76.0 & 74.1 & 92.5 & 89.1 & 86.4 & \\
    \cline{1-7}
    \hline
    \multicolumn{1}{|c||}{SAVER} & \textbf{79.9} & \textbf{77.6} & \textbf{75.3} & \textbf{95.2} & \textbf{91.1} & \textbf{88.3} & \multicolumn{1}{c|}{$\alpha=0.97$} \\
    \hline
    \end{tabular}
    }
    \label{tab:vehicleid_results}
\end{table*}
\begin{table*}
    \centering
    \caption{Performance Comparison on VERI-Wild}
    \resizebox{0.8\columnwidth}{!}{
    \begin{tabular}{c|c|c|c|c|c|c|c|c|c|c}
    \cline{1-10}
    \multicolumn{1}{|c||}{\multirow{2}{*}{Model}} & \multicolumn{3}{c|}{mAP(\%)} & \multicolumn{3}{|c|}{CMC@1(\%)} & \multicolumn{3}{|c|}{CMC@5(\%)} & \\
    \cline{2-10}
    \multicolumn{1}{|c||}{} & S & M & L & S & M & L & S & M & L &\\
    \cline{1-10}
    \multicolumn{1}{|c||}{Baseline} & 78.5 & 72.8 & 65.0 & 92.9 & 91.3 & 88.1 & 97.3 & 96.8 & 95.0 & \\
    \cline{1-10}
    \hline
    \multicolumn{1}{|c||}{SAVER} & \textbf{80.9} & \textbf{75.3} & \textbf{67.7} & \textbf{94.5} & \textbf{92.7} & \textbf{89.5} & \textbf{98.1} & \textbf{97.4} & \textbf{95.8} & \multicolumn{1}{c|}{$\alpha=0.94$} \\
    \hline
    \end{tabular}
    }
    \label{tab:veriwild_results}
\end{table*}
\begin{table*}
    \centering
    \caption{Performance Comparison on Vehicle-1M}
    \resizebox{0.65\columnwidth}{!}{
    \begin{tabular}{c|c|c|c|c|c|c|c}
    \cline{1-7}
    \multicolumn{1}{|c||}{\multirow{2}{*}{Model}} &  \multicolumn{3}{|c|}{CMC@1(\%)} & \multicolumn{3}{|c|}{CMC@5(\%)} & \\
    \cline{2-7}
    \multicolumn{1}{|c||}{} & S & M & L & S & M & L &\\
    \cline{1-7}
    \multicolumn{1}{|c||}{Baseline} & 93.6 & 94.9 & 91.7 & 97.9 & 99.1 & 98.0 & \\
    \cline{1-7}
    \hline
    \multicolumn{1}{|c||}{SAVER} & \textbf{95.5} & \textbf{95.3} & \textbf{93.1} & \textbf{98.0} & \textbf{99.4} & \textbf{98.6} & \multicolumn{1}{c|}{$\alpha=0.98$} \\
    \hline
    \end{tabular}
    }
    \label{tab:vehicle1m_results}
\end{table*}
\begin{table*}
    \centering
    \caption{Performance Comparison on VD1$\&$VD2}
    \resizebox{0.75\columnwidth}{!}{
    \begin{tabular}{c|c|c|c|c|c}
         \cline{1-5}
         \multicolumn{1}{|c||}{Dataset} & \multicolumn{1}{c||}{Model} & mAP(\%) & CMC@1(\%) & CMC@5(\%) & \\
         \cline{1-5}
         \multicolumn{1}{|c||}{\multirow{2}{*}{VD1}} & \multicolumn{1}{c||}{Baseline} & 96.4 & 96.2 & 98.9 &  \\
         \cline{2-6}
         \multicolumn{1}{|c||}{} & \multicolumn{1}{c||}{SAVER} & \textbf{96.7} & \textbf{96.5} & \textbf{99.1} & \multicolumn{1}{c|}{$\alpha=0.99$} \\
         \hline \cline{1-6} 
         \multicolumn{1}{|c||}{\multirow{2}{*}{VD2}} & \multicolumn{1}{c||}{Baseline} & \textbf{96.8} & \textbf{97.9} & 99.0 &  \\
         \cline{2-6}
         \multicolumn{1}{|c||}{} & \multicolumn{1}{c||}{SAVER} & 96.7 & 97.8 & 99.0 & \multicolumn{1}{c|}{$\alpha=0.99$} \\
         \cline{1-6}
    \end{tabular}
    }
    \label{tab:VD}
\end{table*}
\subsubsection{State-of-the-Art Comparison}
In this section, we present the latest state-of-the-art vehicle re-id methods and highlight the performance of the proposed SAVER model. Table \ref{tab:sota_veri_vehicleid} reports the state-of-the-art on re-id benchmarks. It can be seen that our proposed model, despite its simplicity, surpasses the most recent state-of-the-art vehicle re-id works without relying on any extra annotations or attributes. For the case of VeRi and VERI-Wild datasets, we also try the method of re-ranking suggested in \cite{zhong2017re} and achieved considerable mAP scores of $82.0$ and $84.4$ respectively.
\begin{table*}[h!]
    \centering
    \caption{Comparison with recent state-of-the-arts methods}
    \resizebox{0.7\columnwidth}{!}{
    \begin{tabular}{c|c|c|c|c|c|c|c|c|c}
        \cline{2-10} 
        & \multicolumn{9}{|c|}{Dataset}\\
        \cline{1-10}
        \multicolumn{1}{|c||}{\multirow{4}{*}{Method}} & \multicolumn{3}{|c||}{VeRi} & \multicolumn{6}{|c|}{VehicleID}\\
        \cline{2-10}
        \multicolumn{1}{|c||}{} & \multicolumn{1}{|c|}{\multirow{3}{*}{mAP(\%)}} & \multicolumn{2}{|c||}{\multirow{2}{*}{CMC(\%)}} & \multicolumn{2}{|c|}{S} & \multicolumn{2}{|c|}{M} & \multicolumn{2}{|c|}{L}\\
        \cline{5-10}
        \multicolumn{1}{|c||}{} & \multicolumn{1}{|c|}{} & \multicolumn{2}{|c||}{} & \multicolumn{2}{|c|}{CMC(\%)} & \multicolumn{2}{|c|}{CMC(\%)} & \multicolumn{2}{|c|}{CMC(\%)}\\
        \cline{3-10}
        \multicolumn{1}{|c||}{} & \multicolumn{1}{|c|}{} & \multicolumn{1}{|c|}{@1} & \multicolumn{1}{|c||}{@5} &
        \multicolumn{1}{|c|}{@1} & \multicolumn{1}{|c|}{@5} &
        \multicolumn{1}{|c|}{@1} & \multicolumn{1}{|c|}{@5} &
        \multicolumn{1}{|c|}{@1} & \multicolumn{1}{|c|}{@5}\\
        \cline{1-10}
        \multicolumn{1}{|c||}{AAVER \cite{Khorramshahi_2019_ICCV}} & \multicolumn{1}{|c|}{66.35} & \multicolumn{1}{|c|}{90.17} & \multicolumn{1}{|c||}{94.34} & 
        \multicolumn{1}{|c|}{74.69} & \multicolumn{1}{|c|}{93.82} &
        \multicolumn{1}{|c|}{68.62} & \multicolumn{1}{|c|}{89.95} &
        \multicolumn{1}{|c|}{63.54} & \multicolumn{1}{|c|}{85.64}\\
        \cline{1-10} \cline{1-10}
        \multicolumn{1}{|c||}{CCA \cite{peng2019eliminating}} & \multicolumn{1}{|c|}{68.05} & \multicolumn{1}{|c|}{91.71} & \multicolumn{1}{|c||}{96.90} & 
        \multicolumn{1}{|c|}{75.51} & \multicolumn{1}{|c|}{91.14} &
        \multicolumn{1}{|c|}{73.60} & \multicolumn{1}{|c|}{86.46} &
        \multicolumn{1}{|c|}{70.08} & \multicolumn{1}{|c|}{83.20}\\
        \cline{1-10} \cline{1-10}
        \multicolumn{1}{|c||}{BS \cite{kuma2019vehicle}} & \multicolumn{1}{|c|}{67.55} & \multicolumn{1}{|c|}{90.23} & \multicolumn{1}{|c||}{96.42} & 
        \multicolumn{1}{|c|}{78.80} & \multicolumn{1}{|c|}{\textbf{96.17}} &
        \multicolumn{1}{|c|}{73.41} & \multicolumn{1}{|c|}{\textbf{92.57}} &
        \multicolumn{1}{|c|}{69.33} & \multicolumn{1}{|c|}{\textbf{89.45}}\\
        \cline{1-10} \cline{1-10}
        \multicolumn{1}{|c||}{AGNet \cite{zheng2019attributes}} & \multicolumn{1}{|c|}{71.59} & \multicolumn{1}{|c|}{95.61} & \multicolumn{1}{|c||}{96.56} & 
        \multicolumn{1}{|c|}{71.15} & \multicolumn{1}{|c|}{83.78} &
        \multicolumn{1}{|c|}{69.23} & \multicolumn{1}{|c|}{81.41} &
        \multicolumn{1}{|c|}{65.74} & \multicolumn{1}{|c|}{78.28}\\
        \cline{1-10} \cline{1-10}
        \multicolumn{1}{|c||}{VehicleX \cite{yao2019simulating}} & \multicolumn{1}{|c|}{73.26} & \multicolumn{1}{|c|}{94.99} & \multicolumn{1}{|c||}{97.97} & 
        \multicolumn{1}{|c|}{79.81} & \multicolumn{1}{|c|}{93.17} &
        \multicolumn{1}{|c|}{76.74} & \multicolumn{1}{|c|}{90.34} &
        \multicolumn{1}{|c|}{73.88} & \multicolumn{1}{|c|}{88.18}\\
        \cline{1-10} \cline{1-10}
        \multicolumn{1}{|c||}{PRND\cite{he2019part}} & \multicolumn{1}{|c|}{74.3} & \multicolumn{1}{|c|}{94.3} & \multicolumn{1}{|c||}{\textbf{98.7}} & 
        \multicolumn{1}{|c|}{78.4} & \multicolumn{1}{|c|}{92.3} &
        \multicolumn{1}{|c|}{75.0} & \multicolumn{1}{|c|}{88.3} &
        \multicolumn{1}{|c|}{74.2} & \multicolumn{1}{|c|}{86.4}\\
        \hline \hline
        \multicolumn{1}{|c||}{Ours} & \multicolumn{1}{|c|}{\textbf{79.6}} & \multicolumn{1}{|c|}{\textbf{96.4}} & \multicolumn{1}{|c||}{98.6} & 
        \multicolumn{1}{|c|}{\textbf{79.9}} & \multicolumn{1}{|c|}{95.2} &
        \multicolumn{1}{|c|}{\textbf{77.6}} & \multicolumn{1}{|c|}{91.1} &
        \multicolumn{1}{|c|}{\textbf{75.3}} & \multicolumn{1}{|c|}{88.3}\\
        \hline
        \cline{1-4} \cline{1-4} \cline{1-4}
        \multicolumn{1}{|c||}{Ours + Re-ranking} & \multicolumn{1}{|c|}{\textbf{82.0}} & \multicolumn{1}{|c|}{\textbf{96.9}} & \multicolumn{1}{|c||}{97.7} & 
        \multicolumn{1}{c}{} & \multicolumn{1}{c}{} & \multicolumn{1}{c}{} & \multicolumn{1}{c}{} &
        \multicolumn{1}{c}{}& \\
        \cline{1-4}
        \multicolumn{1}{c}{}& \multicolumn{1}{c}{}& \multicolumn{1}{c}{}& \multicolumn{1}{c}{}& \multicolumn{1}{c}{}& \multicolumn{1}{c}{}& \multicolumn{1}{c}{}& \multicolumn{1}{c}{}& \multicolumn{1}{c}{}& \multicolumn{1}{c}{}\\
        \multicolumn{1}{c}{}& \multicolumn{1}{c}{}& \multicolumn{1}{c}{}& \multicolumn{1}{c}{}& \multicolumn{1}{c}{}& \multicolumn{1}{c}{}& \multicolumn{1}{c}{}& \multicolumn{1}{c}{}& \multicolumn{1}{c}{}& \multicolumn{1}{c}{}
        \\
    \end{tabular}
    }
    \resizebox{\columnwidth}{!}{%
    \begin{tabular}{c|c|c|c|c|c|c|c|c|c|c|c|c|c|c|c|c|c|c|c|c|c}
    \cline{2-22} 
    & \multicolumn{21}{|c|}{Dataset}\\
    \cline{1-22}
    \multicolumn{1}{|c||}{\multirow{4}{*}{Method}} & \multicolumn{9}{|c||}{VERI-Wild} & \multicolumn{6}{|c||}{Vehicle-1M} & \multicolumn{3}{|c||}{VD1} & \multicolumn{3}{|c|}{VD2}\\
    \cline{2-22}
    \multicolumn{1}{|c||}{} & \multicolumn{3}{|c|}{S} & \multicolumn{3}{|c|}{M} & \multicolumn{3}{|c||}{L} & \multicolumn{2}{|c|}{S} & \multicolumn{2}{|c|}{M} & \multicolumn{2}{|c||}{L} & \multicolumn{1}{|c|}{\multirow{3}{*}{mAP}} & \multicolumn{2}{|c||}{\multirow{2}{*}{CMC}} & \multicolumn{1}{|c|}{\multirow{3}{*}{mAP}} & \multicolumn{2}{|c|}{\multirow{2}{*}{CMC}}\\
    \cline{2-16}
    \multicolumn{1}{|c||}{} & \multicolumn{1}{|c|}{\multirow{2}{*}{mAP}} & \multicolumn{2}{|c|}{CMC} & \multicolumn{1}{|c|}{\multirow{2}{*}{mAP}} & \multicolumn{2}{|c|}{CMC} &  \multicolumn{1}{|c|}{\multirow{2}{*}{mAP}} & \multicolumn{2}{|c||}{CMC} & 
    \multicolumn{2}{|c|}{CMC} & 
    \multicolumn{2}{|c|}{CMC} & 
    \multicolumn{2}{|c||}{CMC} & \multicolumn{1}{|c|}{} & \multicolumn{1}{|c}{} & \multicolumn{1}{c||}{} & \multicolumn{1}{|c|}{} & \multicolumn{1}{|c}{}& \multicolumn{1}{c|}{}\\
    \cline{18-19} \cline{21-22} \cline{3-4} \cline{6-7} \cline{9-16}
    \multicolumn{1}{|c||}{} & \multicolumn{1}{|c|}{} & \multicolumn{1}{|c|}{@1} & \multicolumn{1}{|c|}{@5} & \multicolumn{1}{|c|}{} & \multicolumn{1}{|c|}{@1} & \multicolumn{1}{|c|}{@5} & \multicolumn{1}{|c|}{} & \multicolumn{1}{|c|}{@1} & \multicolumn{1}{|c||}{@5} &
    \multicolumn{1}{|c|}{@1} & \multicolumn{1}{|c|}{@5} & \multicolumn{1}{|c|}{@1} & \multicolumn{1}{|c|}{@5} & \multicolumn{1}{|c|}{@1} & \multicolumn{1}{|c||}{@5} & \multicolumn{1}{|c|}{} & \multicolumn{1}{|c|}{@1} & \multicolumn{1}{|c||}{@5} & \multicolumn{1}{|c|}{} & \multicolumn{1}{|c|}{@1} & \multicolumn{1}{|c|}{@5}\\
    \cline{1-22} \cline{1-22}
    \multicolumn{1}{|c||}{BS\cite{kuma2019vehicle}} & 
    \multicolumn{1}{|c|}{70.54} & \multicolumn{1}{|c|}{84.17} &  \multicolumn{1}{|c|}{95.30} &  \multicolumn{1}{|c|}{62.83} &  \multicolumn{1}{|c|}{78.22} &  \multicolumn{1}{|c|}{93.06} &  \multicolumn{1}{|c|}{51.63} & \multicolumn{1}{|c|}{69.99} &  \multicolumn{1}{|c||}{88.45} & \multicolumn{1}{|c|}{-} & \multicolumn{1}{|c|}{-} & \multicolumn{1}{|c|}{-} & \multicolumn{1}{|c|}{-} & \multicolumn{1}{|c|}{-} & \multicolumn{1}{|c||}{-} & \multicolumn{1}{|c|}{87.48} & \multicolumn{1}{|c|}{-} &  \multicolumn{1}{|c||}{-} &\multicolumn{1}{|c|}{84.55} & \multicolumn{1}{|c|}{-} &  \multicolumn{1}{|c|}{-}\\
    \cline{1-22} \cline{1-22}
    \multicolumn{1}{|c||}{AAVER\cite{Khorramshahi_2019_ICCV}} & 
    \multicolumn{1}{|c|}{62.23} & \multicolumn{1}{|c|}{75.80} &  \multicolumn{1}{|c|}{92.70} &  \multicolumn{1}{|c|}{53.66} &  \multicolumn{1}{|c|}{68.24} &  \multicolumn{1}{|c|}{88.88} &  \multicolumn{1}{|c|}{41.68} & \multicolumn{1}{|c|}{58.69} &  \multicolumn{1}{|c||}{81.59} & \multicolumn{1}{|c|}{-} & \multicolumn{1}{|c|}{-} & \multicolumn{1}{|c|}{-} & \multicolumn{1}{|c|}{-} & \multicolumn{1}{|c|}{-} & \multicolumn{1}{|c||}{-} & \multicolumn{1}{|c|}{-} & \multicolumn{1}{|c|}{-} &  \multicolumn{1}{|c||}{-} &\multicolumn{1}{|c|}{-} & \multicolumn{1}{|c|}{-} &  \multicolumn{1}{|c|}{-}\\
    \cline{1-22} \cline{1-22}
    \multicolumn{1}{|c||}{TAMR \cite{guo2019two}} & 
    \multicolumn{1}{|c|}{-} & \multicolumn{1}{|c|}{-} &  \multicolumn{1}{|c|}{-} &  \multicolumn{1}{|c|}{-} &  \multicolumn{1}{|c|}{-} &  \multicolumn{1}{|c|}{-} &  \multicolumn{1}{|c|}{-} & \multicolumn{1}{|c|}{-} &  \multicolumn{1}{|c||}{-} & \multicolumn{1}{|c|}{\textbf{95.95}} & \multicolumn{1}{|c|}{\textbf{99.24}} & \multicolumn{1}{|c|}{94.27} & \multicolumn{1}{|c|}{98.86} & \multicolumn{1}{|c|}{92.91} & \multicolumn{1}{|c||}{98.30} & \multicolumn{1}{|c|}{-} & \multicolumn{1}{|c|}{-} &  \multicolumn{1}{|c||}{-} &\multicolumn{1}{|c|}{-} & \multicolumn{1}{|c|}{-} &  \multicolumn{1}{|c|}{-}\\
    \hline \hline
    \multicolumn{1}{|c||}{Ours} & 
    \multicolumn{1}{|c|}{\textbf{80.9}} & \multicolumn{1}{|c|}{\textbf{94.5}} &  \multicolumn{1}{|c|}{\textbf{98.1}} &  \multicolumn{1}{|c|}{\textbf{75.3}} &  \multicolumn{1}{|c|}{\textbf{92.7}} &  \multicolumn{1}{|c|}{\textbf{97.4}} &  \multicolumn{1}{|c|}{\textbf{67.7}} & \multicolumn{1}{|c|}{\textbf{89.5}} &  \multicolumn{1}{|c||}{\textbf{95.8}} & \multicolumn{1}{|c|}{95.5} & \multicolumn{1}{|c|}{98.0} & \multicolumn{1}{|c|}{\textbf{95.3}} & \multicolumn{1}{|c|}{\textbf{99.4}} & \multicolumn{1}{|c|}{\textbf{93.1}} & \multicolumn{1}{|c||}{\textbf{98.6}} & \multicolumn{1}{|c|}{\textbf{96.7}} & \multicolumn{1}{|c|}{\textbf{96.5}} &  \multicolumn{1}{|c||}{\textbf{99.1}} &\multicolumn{1}{|c|}{\textbf{96.7}} & \multicolumn{1}{|c|}{\textbf{97.8}} &  \multicolumn{1}{|c|}{\textbf{99.0}}\\
    \hline \cline{1-4} \cline{1-4} \cline{1-4}
    \multicolumn{1}{|c||}{Ours + } & 
    \multicolumn{1}{|c|}{\multirow{2}{*}{\textbf{84.4}}} & \multicolumn{1}{|c|}{\multirow{2}{*}{\textbf{95.3}}} &  \multicolumn{1}{|c|}{\multirow{2}{*}{97.6}} & \multicolumn{1}{c}{} & \multicolumn{1}{c}{}& \multicolumn{1}{c}{}& \multicolumn{1}{c}{}& \multicolumn{1}{c}{}& \multicolumn{1}{c}{}& \multicolumn{1}{c}{}& \multicolumn{1}{c}{}& \multicolumn{1}{c}{}& \multicolumn{1}{c}{}& \multicolumn{1}{c}{}& \multicolumn{1}{c}{}& \multicolumn{1}{c}{}& \multicolumn{1}{c}{}& \multicolumn{1}{c}{}& \multicolumn{1}{c}{}& \multicolumn{1}{c}{}& \multicolumn{1}{c}{} \\
    \multicolumn{1}{|c||}{Re-ranking} & \multicolumn{1}{|c|}{} & \multicolumn{1}{|c|}{} & \multicolumn{1}{|c|}{}\\ 
    \cline{1-4}
    \end{tabular}
    }
    \label{tab:sota_veri_vehicleid}
\end{table*}

\section{Ablation Studies}
\label{sec:ablation}
In this section, we design a set of experiments to study the impact of different neural network architectures on the quality of reconstructed images, and also understand the impact of key hyper-parameters. In addition, we are interested in understanding how we can maximally exploit the reconstructed images in deep feature extraction. The experimental results of the reconstruction network are evaluated on the Vehicle Universe dataset, and experiments regarding the deep feature extraction module are evaluated on VeRi and VehicleID datasets.

\subsection{Residual Generation Techniques}
\begin{figure}
    \centering
    \subfloat[][Original]{\includegraphics[width=0.15\textwidth, height=0.15\textwidth]{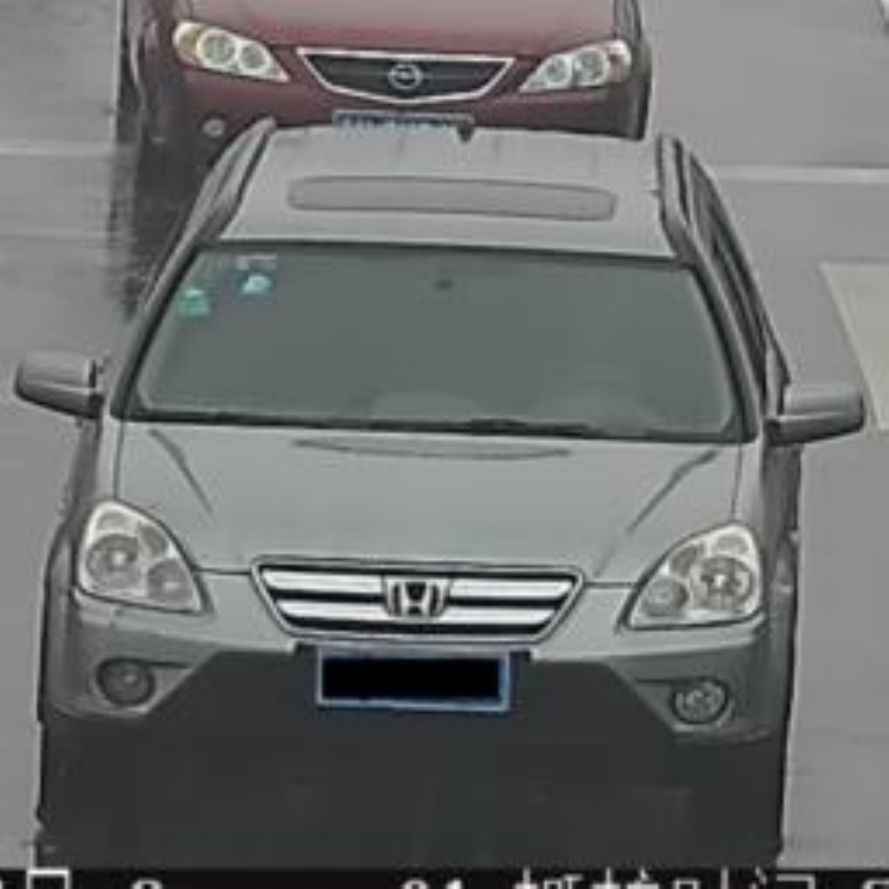}}\quad
    \subfloat[][AE]{\includegraphics[width=0.15\textwidth, height=0.15\textwidth]{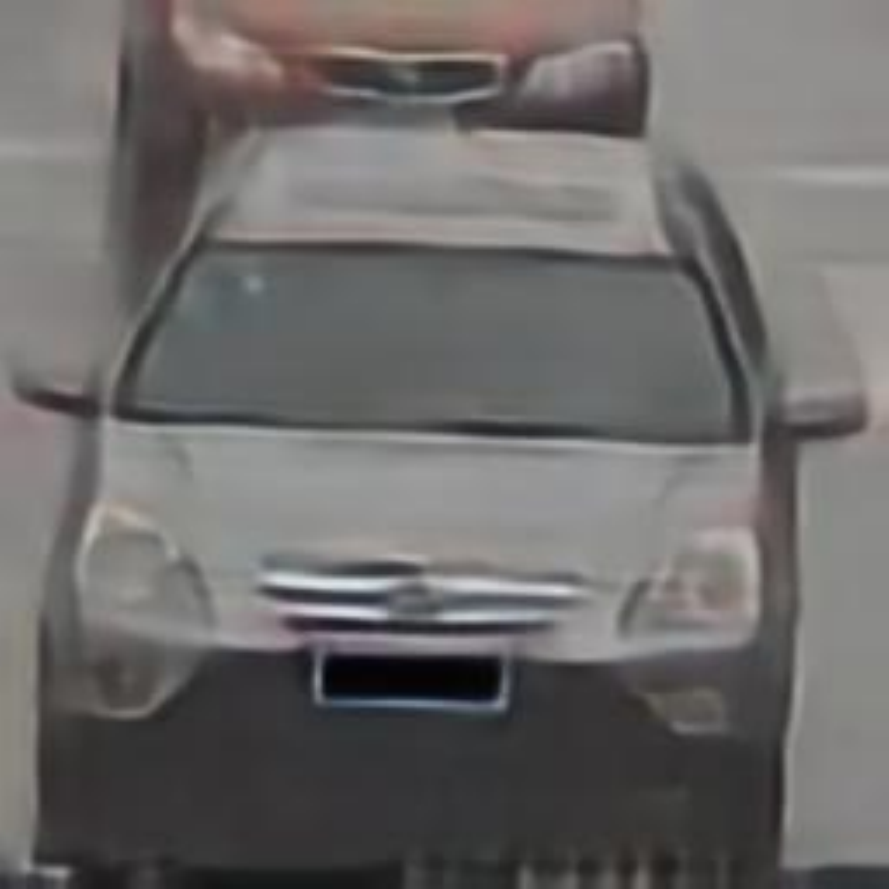}}\quad
    \subfloat[][VAE]{\includegraphics[width=0.15\textwidth, height=0.15\textwidth]{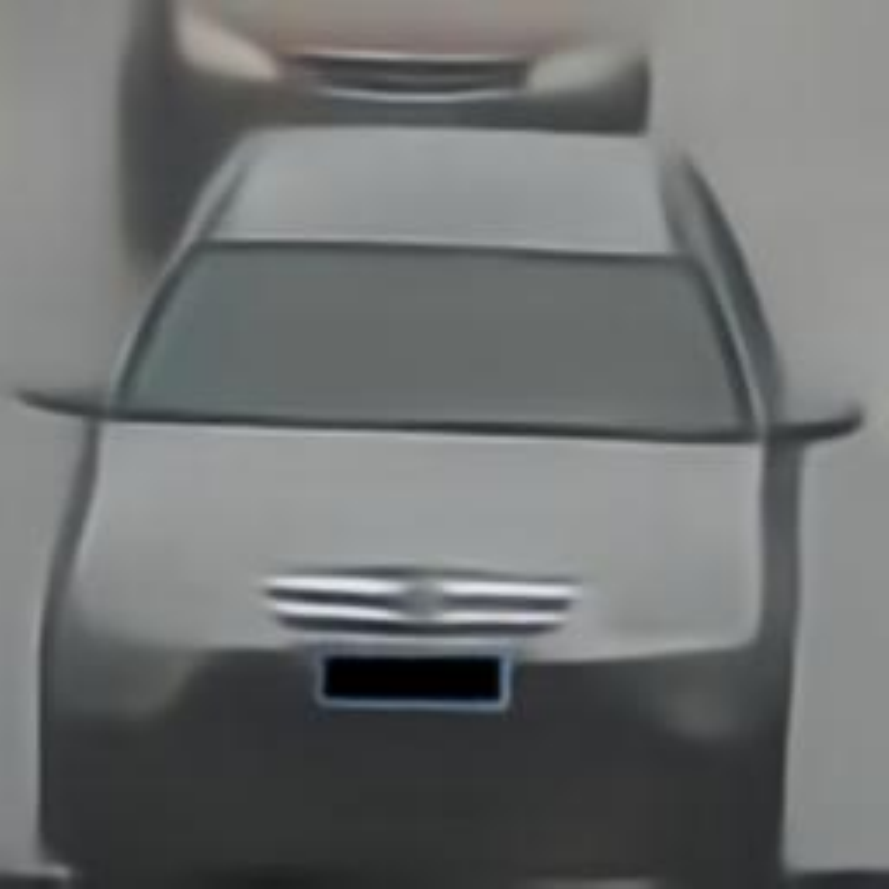}}\quad
    \subfloat[][GAN]{\includegraphics[width=0.15\textwidth, height=0.15\textwidth]{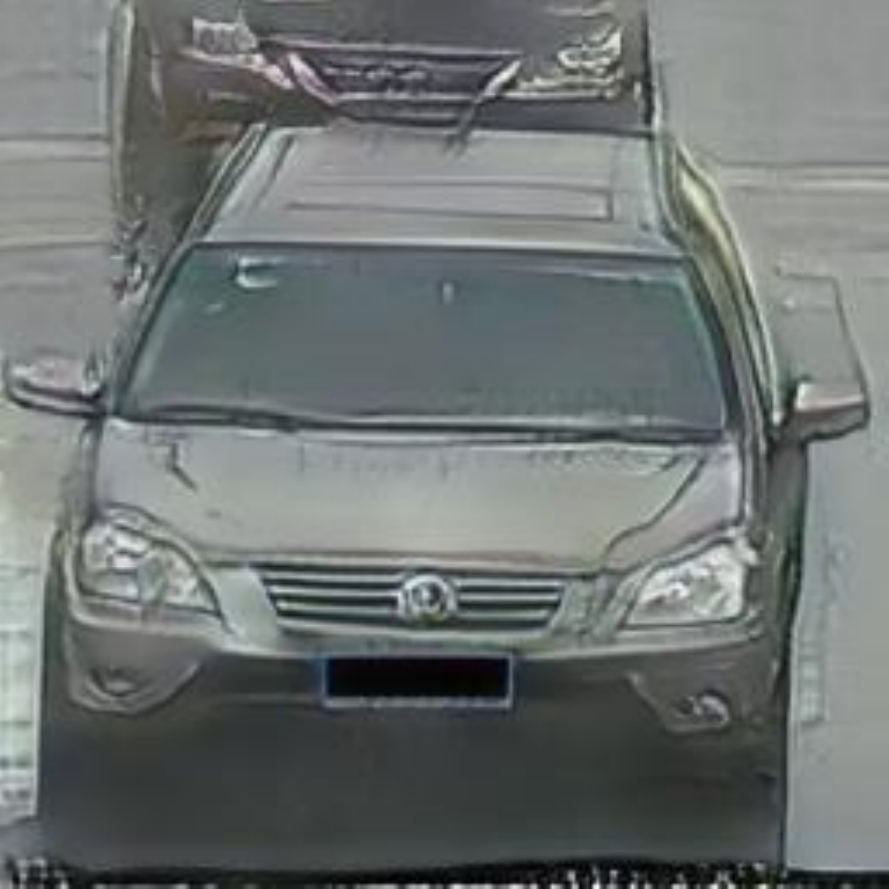}}
    \quad
    \subfloat[][BF]{\includegraphics[width=0.15\textwidth, height=0.15\textwidth]{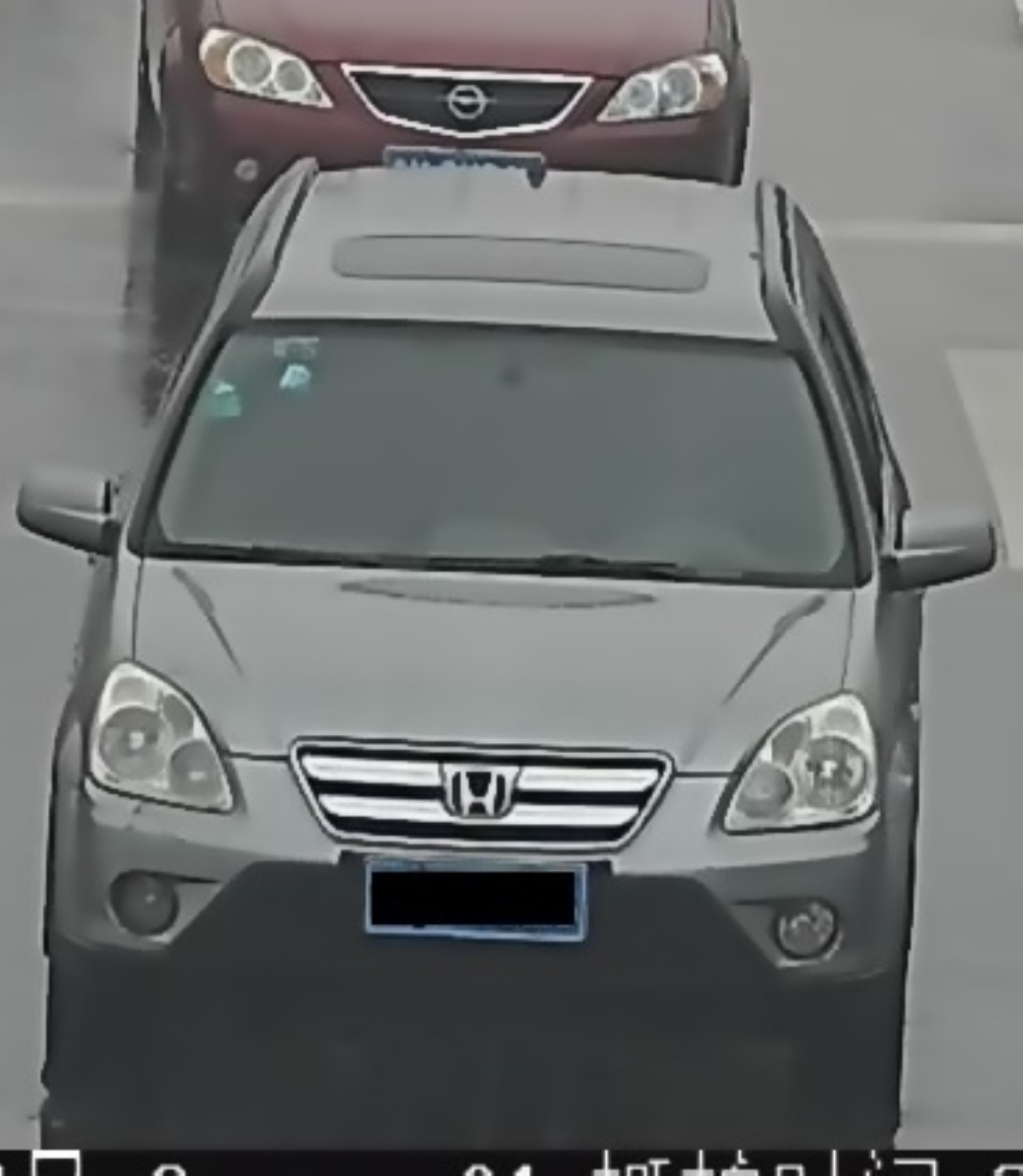}}
    \\
    \subfloat[][Original]{\includegraphics[width=0.15\textwidth, height=0.15\textwidth]{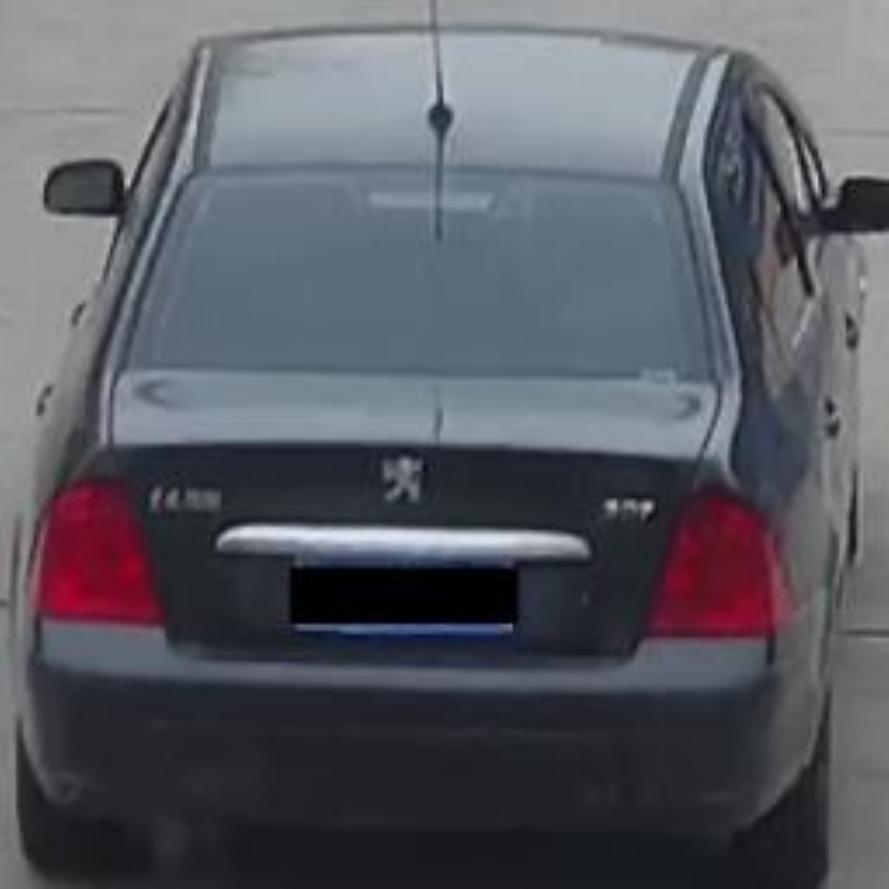}}\quad
    \subfloat[][AE]{\includegraphics[width=0.15\textwidth, height=0.15\textwidth]{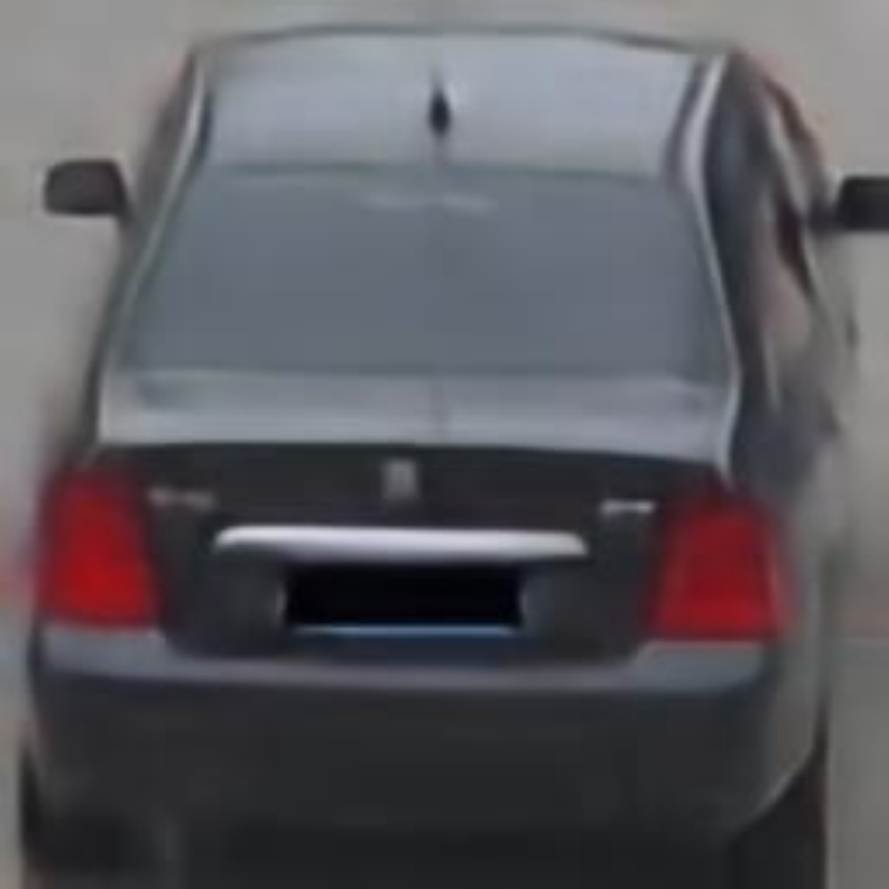}}\quad
    \subfloat[][VAE]{\includegraphics[width=0.15\textwidth, height=0.15\textwidth]{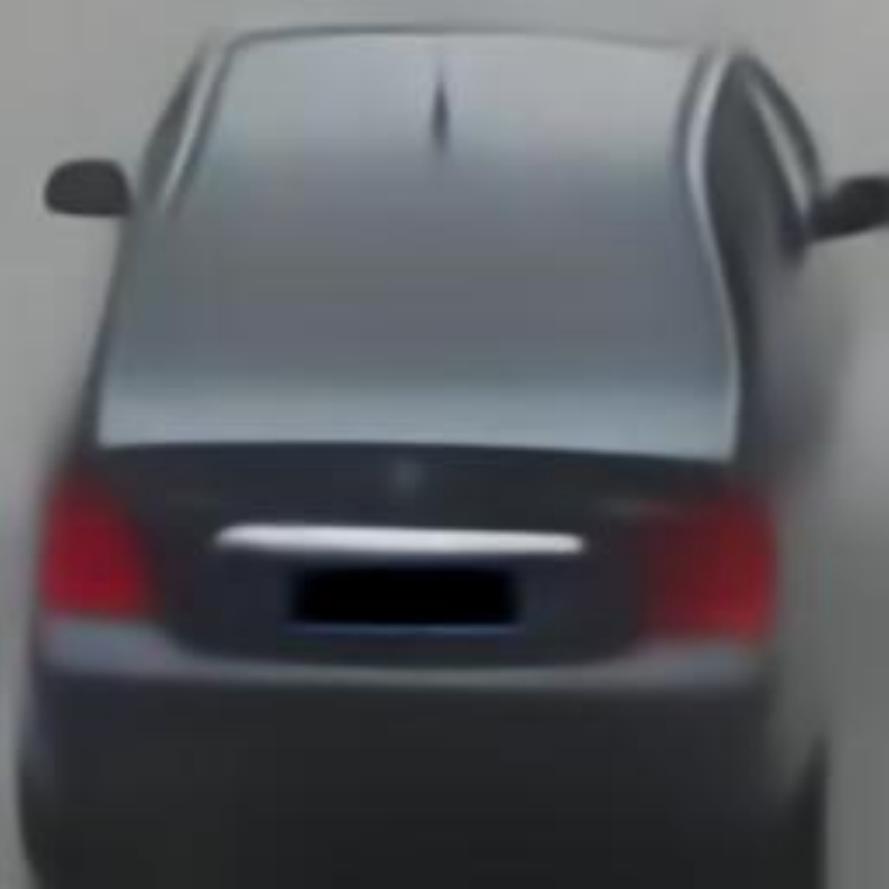}}\quad
    \subfloat[][GAN]{\includegraphics[width=0.15\textwidth, height=0.15\textwidth]{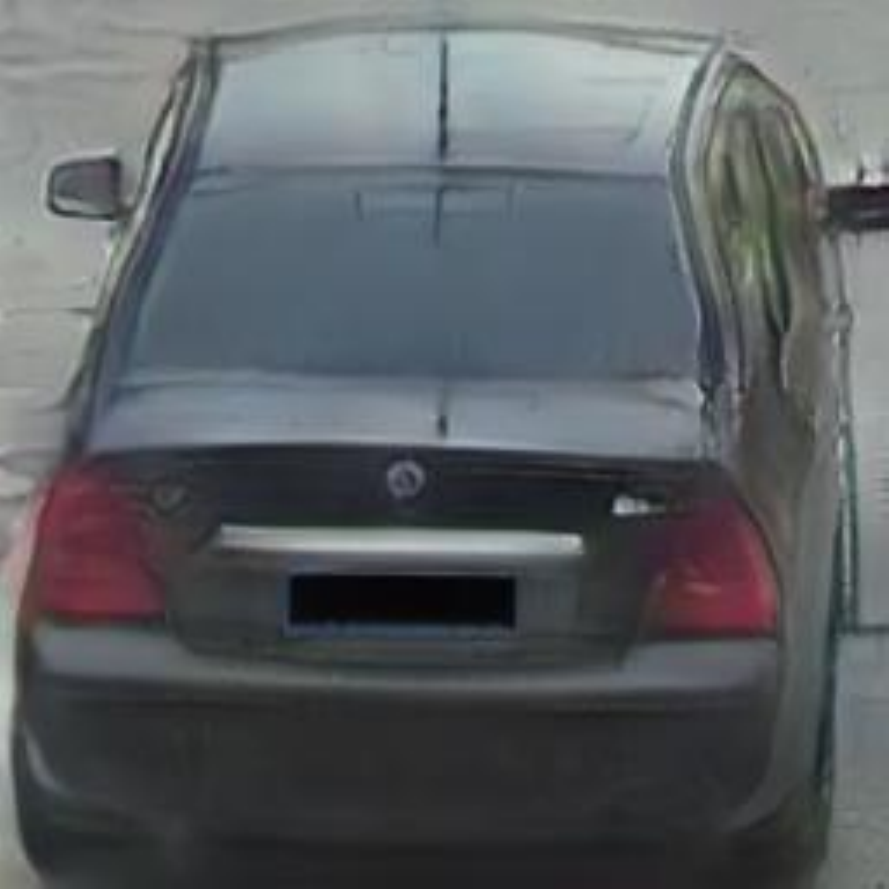}}\quad
    \subfloat[][\small{BF}]{\includegraphics[width=0.15\textwidth, height=0.15\textwidth]{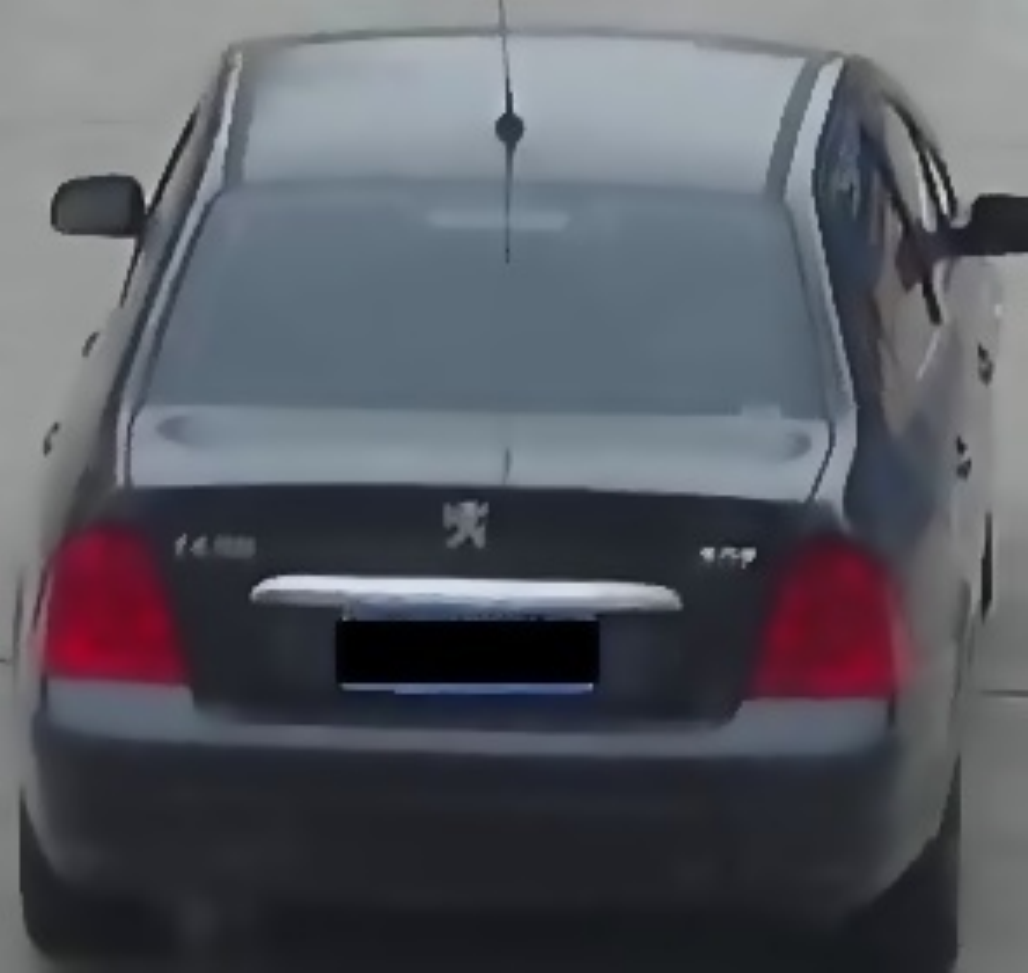}}
    \caption{Different image reconstruction methods.}
    \label{fig:ablation_res_generation}
\end{figure}
\subsubsection{Effect of Different Reconstruction Architectures}
Here, we study the reconstruction quality of Auto-Encoder (AE) \cite{BadrinarayananSegNet}, VAE \cite{kingmaVAE}, and GAN \cite{goodfellowGAN} methods. Moreover, we study the use of Bilateral Filtering (BF) as a baseline for texture smoothing, subsequent residual generation and vehicle re-id. Figure \ref{fig:ablation_res_generation} qualitatively illustrates the reconstruction of each method for a given vehicle identity. We notice that both AE and GAN models attempt to recreate fine-grained details, but often introduce additional distortions. Specifically, the GAN model generates new textures, modifies the logo and distorts the overall shape of the vehicle. As a result, GANs produce sharper images with various artifacts that diminish the quality of the residual image required by the re-id network. Also note that although bilateral filtering attempts to smooth images, it is unable to remove the critical details needed in residuals and vehicle re-id. The VAE is able to reconstruct the image by removing minute details and smoothing out textures. As a result, the VAE is able to generate the detailed residual maps needed for our proposed re-id method. Table \ref{tab:ablation_reconstruction_methods} presents evaluation metrics on VeRi-766 and VehicleID for each of the generative models and bilateral filtering.

\begin{table*}
    \centering
    \caption{Performance comparison of different image reconstruction methods}
    \resizebox{0.7\columnwidth}{!}{
    \begin{tabular}{c|c|c|c|c|c|c|c|c|c}
        \cline{2-10} 
        & \multicolumn{9}{|c|}{Dataset}\\
        \cline{1-10}
        \multicolumn{1}{|c||}{\multirow{4}{*}{Method}} & \multicolumn{3}{|c||}{VeRi} & \multicolumn{6}{|c|}{VehicleID}\\
        \cline{2-10}
        \multicolumn{1}{|c||}{} & \multicolumn{1}{|c|}{\multirow{3}{*}{mAP(\%)}} & \multicolumn{2}{|c||}{\multirow{2}{*}{CMC(\%)}} & \multicolumn{2}{|c|}{S} & \multicolumn{2}{|c|}{M} & \multicolumn{2}{|c|}{L}\\
        \cline{5-10}
        \multicolumn{1}{|c||}{} & \multicolumn{1}{|c|}{} & \multicolumn{2}{|c||}{} & \multicolumn{2}{|c|}{CMC(\%)} & \multicolumn{2}{|c|}{CMC(\%)} & \multicolumn{2}{|c|}{CMC(\%)}\\
        \cline{3-10}
        \multicolumn{1}{|c||}{} & \multicolumn{1}{|c|}{} & \multicolumn{1}{|c|}{@1} & \multicolumn{1}{|c||}{@5} &
        \multicolumn{1}{|c|}{@1} & \multicolumn{1}{|c|}{@5} &
        \multicolumn{1}{|c|}{@1} & \multicolumn{1}{|c|}{@5} &
        \multicolumn{1}{|c|}{@1} & \multicolumn{1}{|c|}{@5}\\
        \cline{1-10}
        \multicolumn{1}{|c||}{AE} & \multicolumn{1}{|c|}{79.0} & \multicolumn{1}{|c|}{96.0} & \multicolumn{1}{|c||}{98.2} & 
        \multicolumn{1}{|c|}{79.0} & \multicolumn{1}{|c|}{93.9} &
        \multicolumn{1}{|c|}{76.8} & \multicolumn{1}{|c|}{90.5} &
        \multicolumn{1}{|c|}{74.9} & \multicolumn{1}{|c|}{87.9}\\
        \cline{1-10} \cline{1-10}
        \multicolumn{1}{|c||}{VAE} & \multicolumn{1}{|c|}{\textbf{79.6}} & \multicolumn{1}{|c|}{\textbf{96.4}} & \multicolumn{1}{|c||}{\textbf{98.6}} & 
        \multicolumn{1}{|c|}{\textbf{79.9}} & \multicolumn{1}{|c|}{\textbf{95.2}} &
        \multicolumn{1}{|c|}{\textbf{77.6}} & \multicolumn{1}{|c|}{\textbf{91.1}} &
        \multicolumn{1}{|c|}{\textbf{75.3}} & \multicolumn{1}{|c|}{\textbf{88.3}}\\
        \cline{1-10} \cline{1-10}
        \multicolumn{1}{|c||}{GAN} & \multicolumn{1}{|c|}{78.3} & \multicolumn{1}{|c|}{95.6} & \multicolumn{1}{|c||}{98.1} & 
        \multicolumn{1}{|c|}{78.5} & \multicolumn{1}{|c|}{93.0} &
        \multicolumn{1}{|c|}{75.6} & \multicolumn{1}{|c|}{89.1} &
        \multicolumn{1}{|c|}{73.4} & \multicolumn{1}{|c|}{85.7}\\
        \cline{1-10} \cline{1-10}
        \multicolumn{1}{|c||}{BF} & \multicolumn{1}{|c|}{78.5} & \multicolumn{1}{|c|}{95.5} & \multicolumn{1}{|c||}{97.6} & 
        \multicolumn{1}{|c|}{78.7} & \multicolumn{1}{|c|}{76.6} &
        \multicolumn{1}{|c|}{74.5} & \multicolumn{1}{|c|}{94.2} &
        \multicolumn{1}{|c|}{90.2} & \multicolumn{1}{|c|}{87.4}\\
        \cline{1-10} \cline{1-10}
    \end{tabular}
    }
    \label{tab:ablation_reconstruction_methods}
\end{table*}
\subsubsection{Effect of Scaling Kullbeck-Leibler Divergence Coefficient $\lambda$ in Eq. \ref{eq:reconstruction_loss}}
\begin{figure}
    \centering
    \subfloat[][Original]{\includegraphics[width=0.18\textwidth, height=0.18\textwidth]{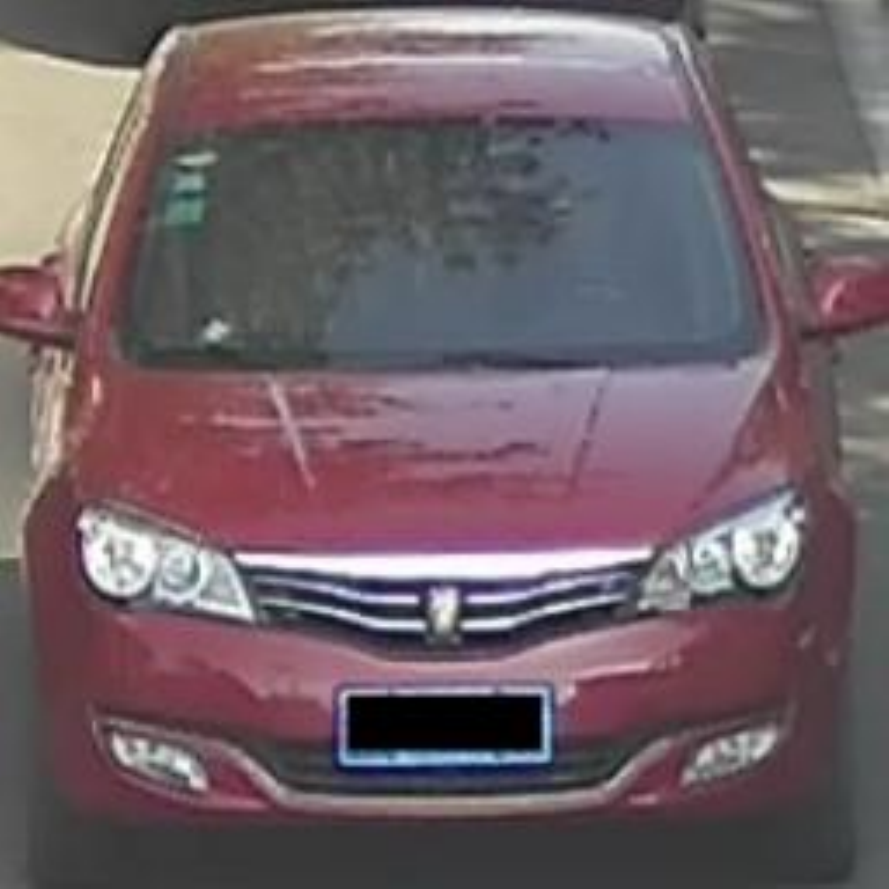}}\quad
    \subfloat[][$\lambda=1e-1$]{\includegraphics[width=0.18\textwidth, height=0.18\textwidth]{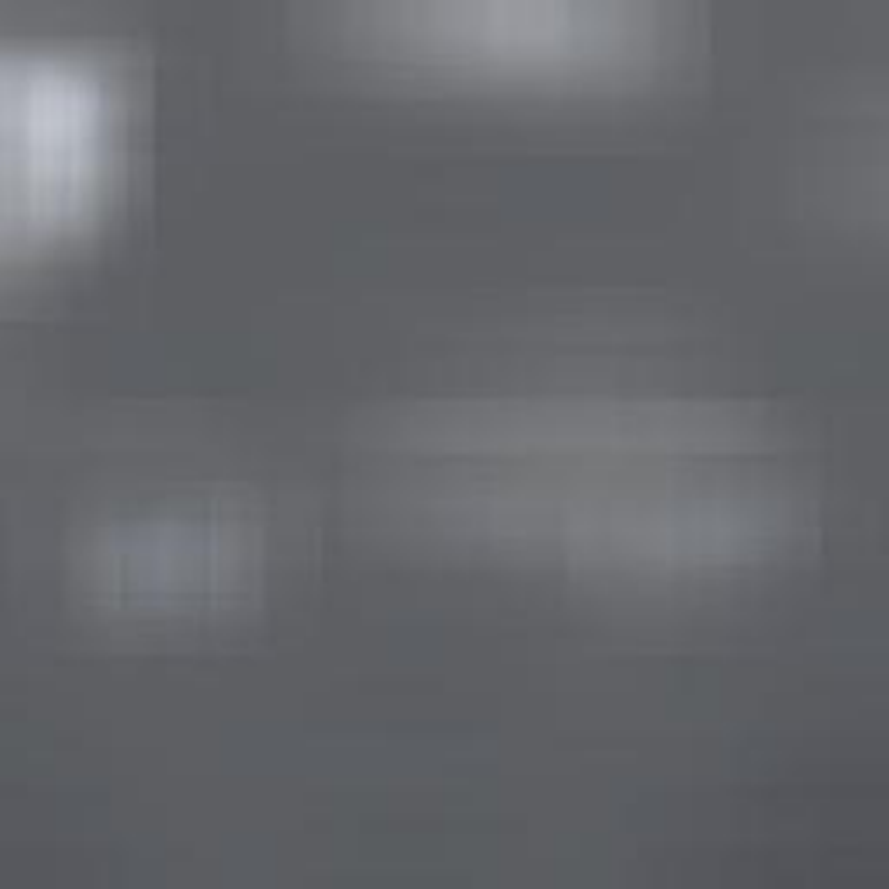}}\quad
    \subfloat[][$\lambda=1e-2$]{\includegraphics[width=0.18\textwidth, height=0.18\textwidth]{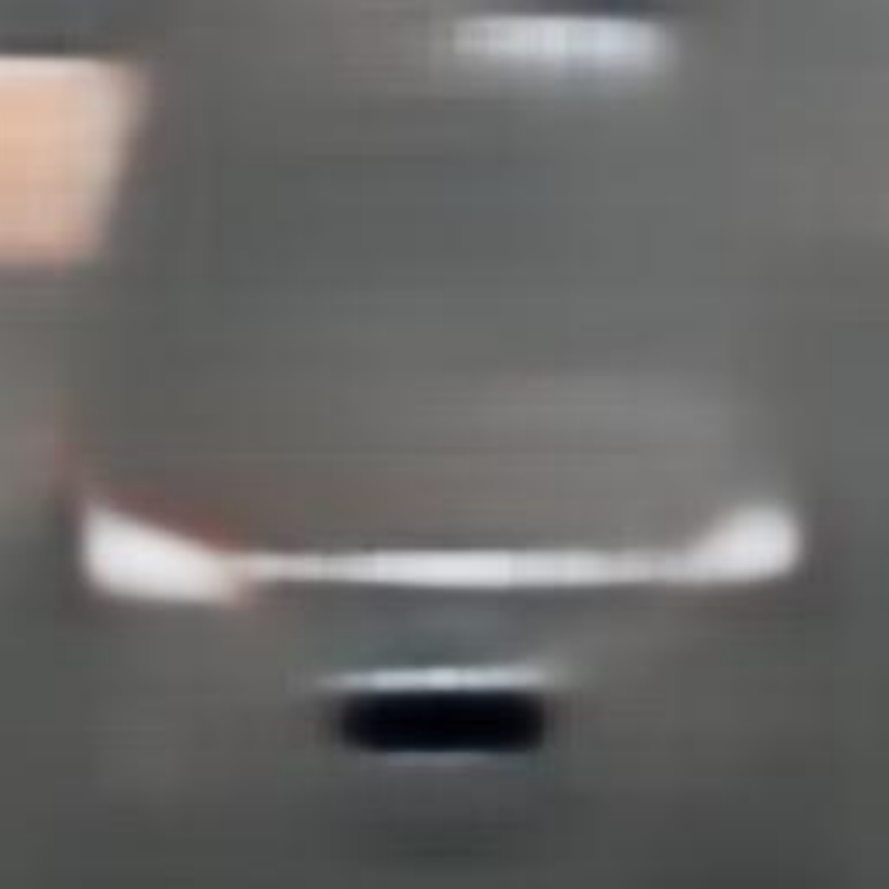}}\quad
    \subfloat[][$\lambda=1e-3$]{\includegraphics[width=0.18\textwidth, height=0.18\textwidth]{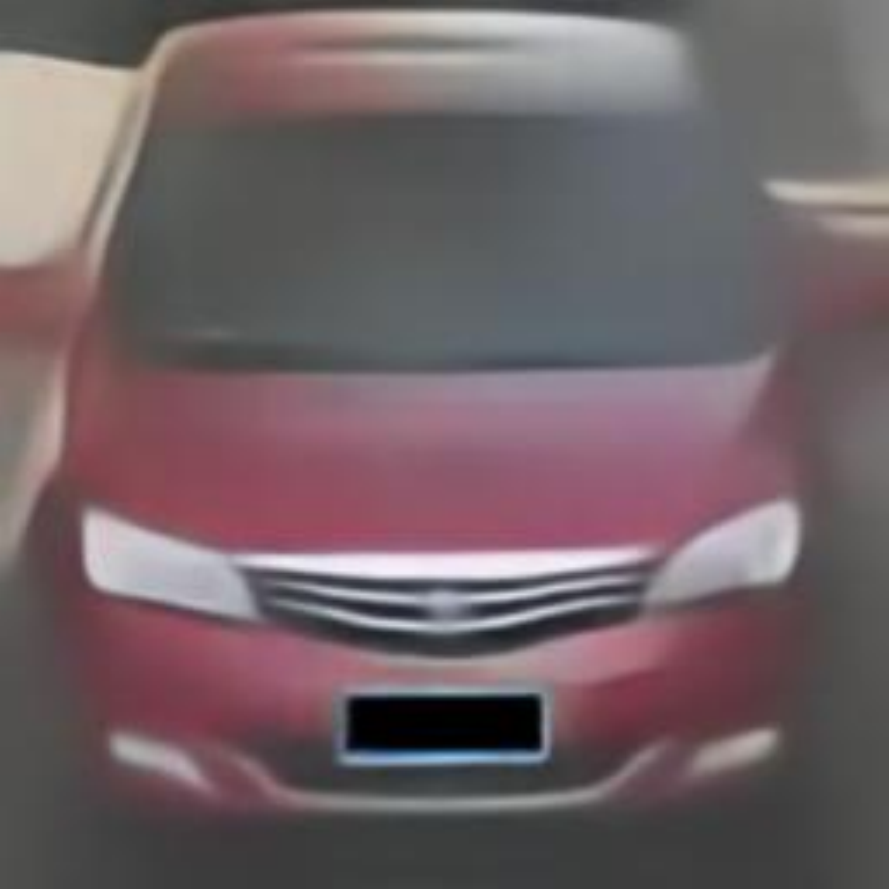}}
    \caption{Effect of scaling KL loss in image generation}
    \label{fig:ablation_KL_scale}
\end{figure}

In this experiment, we are particularly interested in the scaling parameter $\lambda$ used in training the VAE model. Figure \ref{fig:ablation_KL_scale} demonstrates how larger values of $\lambda$ result in a more blurry reconstruction. Intuitively, this parameter offers a natural level for balancing the reconstruction quality of fine-grained discriminative features. As $\lambda$ approaches $0$, our VAE model approximates the reconstruction quality of a traditional Auto-Encoder. Empirically, we found that $\lambda = 1e-3$ produces higher quality vehicle templates, while removing discriminative information across all datasets.

\subsection{Incorporating Residual Information}
To effectively exploit complimentary information provided by the residuals, we design a set of four additional experiments on the VeRi and VehicleID datasets as follows:
\begin{itemize}
    \item[A.] We only feed the VAE reconstruction $I_g$ as input to the re-id network. The purpose of this experiment is to understand how much critical information can be inferred from the VAE reconstruction. 
    \item[B.] We only feed the residual image $I_r$ into the re-id pipeline. In this experiment we are interested to find out how much identity-dependent information can be extracted from only the residual image.
    \item[C.] We use the residual maps to excite the actual image of the vehicle through point-wise matrix multiplication. 
    \item[D.] We concatenate the residual image with the original input image. Therefore, in this experiment we feed a six-channel tensor to the feature extraction module. 
\end{itemize}
Table \ref{tab:ablation_residual_incorporation} presents the results of experiments A to D and highlights their performance against the baseline and SAVER models. In experiment A, the deep feature extractor is trained using the reconstructed image from the VAE. Intuitively, this method provides the lowest performance since all discriminating details are obfuscated. Interestingly, experiment B, training a deep feature extractor using only residual images, is able to perform nearly as well as our standard baseline. This reaffirms the idea that local information is essential for vehicle re-id. Experiment C performs considerably worse than the baseline model, indicating that point-wise multiplication with the sparse residual removes key information. Lastly, experiment D performs lower than our baseline. This can be attributed to the ImageNet \cite{deng2009imagenet} weight initialization, which is not well suited for six-channel images.
\begin{table*}
    \centering
    \caption{Evaluation of different designs of employing residuals}
    \resizebox{0.8\columnwidth}{!}{
    \begin{tabular}{c|c|c|c|c|c|c|c|c|c}
        \cline{2-10} 
        & \multicolumn{9}{|c|}{Dataset}\\
        \cline{1-10}
        \multicolumn{1}{|c||}{\multirow{4}{*}{Experiment}} & \multicolumn{3}{|c||}{VeRi} & \multicolumn{6}{|c|}{VehicleID}\\
        \cline{2-10}
        \multicolumn{1}{|c||}{} & \multicolumn{1}{|c|}{\multirow{3}{*}{mAP(\%)}} & \multicolumn{2}{|c||}{\multirow{2}{*}{CMC(\%)}} & \multicolumn{2}{|c|}{S} & \multicolumn{2}{|c|}{M} & \multicolumn{2}{|c|}{L}\\
        \cline{5-10}
        \multicolumn{1}{|c||}{} & \multicolumn{1}{|c|}{} & \multicolumn{2}{|c||}{} & \multicolumn{2}{|c|}{CMC(\%)} & \multicolumn{2}{|c|}{CMC(\%)} & \multicolumn{2}{|c|}{CMC(\%)}\\
        \cline{3-10}
        \multicolumn{1}{|c||}{} & \multicolumn{1}{|c|}{} & \multicolumn{1}{|c|}{@1} & \multicolumn{1}{|c||}{@5} &
        \multicolumn{1}{|c|}{@1} & \multicolumn{1}{|c|}{@5} &
        \multicolumn{1}{|c|}{@1} & \multicolumn{1}{|c|}{@5} &
        \multicolumn{1}{|c|}{@1} & \multicolumn{1}{|c|}{@5}\\
        \cline{1-10}
        \multicolumn{1}{|c||}{A} & \multicolumn{1}{|c|}{67.5} & \multicolumn{1}{|c|}{91.4} & \multicolumn{1}{|c||}{96.4} & 
        \multicolumn{1}{|c|}{64.2} & \multicolumn{1}{|c|}{80.6} &
        \multicolumn{1}{|c|}{62.9} & \multicolumn{1}{|c|}{76.3} &
        \multicolumn{1}{|c|}{59.4} & \multicolumn{1}{|c|}{73.5}\\
        \cline{1-10} \cline{1-10}
        \multicolumn{1}{|c||}{B} & \multicolumn{1}{|c|}{77.5} & \multicolumn{1}{|c|}{94.5} & \multicolumn{1}{|c||}{98.2} & 
        \multicolumn{1}{|c|}{77.9} & \multicolumn{1}{|c|}{92.7} &
        \multicolumn{1}{|c|}{74.7} & \multicolumn{1}{|c|}{89.0} &
        \multicolumn{1}{|c|}{73.4} & \multicolumn{1}{|c|}{86.2}\\
        \cline{1-10} \cline{1-10}
        \multicolumn{1}{|c||}{C} & \multicolumn{1}{|c|}{71.4} & \multicolumn{1}{|c|}{91.9} & \multicolumn{1}{|c||}{96.4} & 
        \multicolumn{1}{|c|}{76.3} & \multicolumn{1}{|c|}{92.6} &
        \multicolumn{1}{|c|}{73.3} & \multicolumn{1}{|c|}{86.8} &
        \multicolumn{1}{|c|}{70.7} & \multicolumn{1}{|c|}{83.5}\\
        \cline{1-10} \cline{1-10}
        \multicolumn{1}{|c||}{D} & \multicolumn{1}{|c|}{75.7} & \multicolumn{1}{|c|}{94.8} & \multicolumn{1}{|c||}{98.3} & 
        \multicolumn{1}{|c|}{78.9} & \multicolumn{1}{|c|}{93.1} &
        \multicolumn{1}{|c|}{75.3} & \multicolumn{1}{|c|}{89.2} &
        \multicolumn{1}{|c|}{73.3} & \multicolumn{1}{|c|}{86.1}\\
        \hline \hline
        \multicolumn{1}{|c||}{Baseline} & \multicolumn{1}{|c|}{78.2} & \multicolumn{1}{|c|}{95.5} & \multicolumn{1}{|c||}{97.9} & 
        \multicolumn{1}{|c|}{78.4} & \multicolumn{1}{|c|}{92.5} &
        \multicolumn{1}{|c|}{76.0} & \multicolumn{1}{|c|}{89.1} &
        \multicolumn{1}{|c|}{74.1} & \multicolumn{1}{|c|}{86.4}\\
        \cline{1-10} \cline{1-10}
        \multicolumn{1}{|c||}{SAVER} & \multicolumn{1}{|c|}{\textbf{79.6}} & \multicolumn{1}{|c|}{\textbf{96.4}} & \multicolumn{1}{|c||}{\textbf{98.6}} & 
        \multicolumn{1}{|c|}{\textbf{79.9}} & \multicolumn{1}{|c|}{\textbf{95.2}} &
        \multicolumn{1}{|c|}{\textbf{77.6}} & \multicolumn{1}{|c|}{\textbf{91.1}} &
        \multicolumn{1}{|c|}{\textbf{75.3}} & \multicolumn{1}{|c|}{\textbf{88.3}}\\
        \cline{1-10} \cline{1-10}
    \end{tabular}
    }
    \label{tab:ablation_residual_incorporation}
\end{table*}
\section{Conclusion}
In this paper we have shown the benefits of using simple, highly-scalable network architectures and training procedures to generate robust deep features for the task of vehicle re-identification. Our model highlights the importance of attending to discriminative regions without additional annotations, and outperforms existing state-of-the-art methods on benchmark datasets including VeRi, VehicleID, Vehicle-1M, and VeRi-Wild.
\section{Acknowledgement}
This research is supported in part by the Northrop Grumman Mission Systems Research in Applications for Learning Machines (REALM) initiative, and by the Office of the Director of National Intelligence (ODNI), Intelligence Advanced Research Projects Activity (IARPA), via IARPA R$\&$D Contract No. D17PC00345. The views and conclusions contained herein are those of the authors and should not be interpreted as necessarily representing the official policies or endorsements, either expressed or implied, of ODNI, IARPA, or the U.S. Government. The U.S. Government is authorized to reproduce and distribute reprints for Governmental purposes notwithstanding any copyright annotation thereon.

%
%

\end{document}